\documentclass{article}
\usepackage{iclr2027_conference,times}

\usepackage[english]{babel} \exhyphenpenalty=0 
\usepackage{amsmath}
\usepackage{amssymb}

\newcommand{\gP}{\mathcal{P}}              
\newcommand{\gS}{\mathcal{S}}              
\newcommand{\gE}{\mathcal{E}}             
\newcommand{\gG}{\mathcal{G}}             

\newcommand{\matW}{\mathbf{W}}            

\newcommand{\Neighbor}{\mathcal{N}} 

\newcommand{\thresh}{\tau}                 

\usepackage{pifont}
\usepackage{url}
\usepackage{enumitem}
\usepackage{booktabs}
\usepackage{todonotes}
\usepackage{wrapfig}
\usepackage{subcaption}
\usepackage{tcolorbox}
\usepackage{xcolor, colortbl}
\usepackage{multirow}
\usepackage{float}
\usepackage{array}
\usepackage{adjustbox}
\usepackage{etoc}
\usepackage{graphicx}
\usepackage{microtype} 
\usepackage{amsthm}
\usepackage{algorithm}
\usepackage{algpseudocode}
\usepackage{hyperref}
\hypersetup{breaklinks=true,hypertexnames=false}
\theoremstyle{definition}

\newcounter{obs}
\newcommand{\obs}{\refstepcounter{obs}\textbf{Observation \theobs. }}
\tcbuselibrary{skins, breakable}

\title{Corpus-Guided Dual-Path Propagation for Graph Retrieval-Augmented Generation}

\iclrfinalcopy 

\author{
Baoxian Liu\textmd{\textsuperscript{1,3}}\textmd{,} ~Tong Wei\textmd{\textsuperscript{2,3}}\footnotemark[1] \\
\textsuperscript{1}College of Software Engineering, Southeast University, Nanjing 210096, China\\
\textsuperscript{2}School of Computer Science and Engineering, Southeast University, Nanjing 210096, China\\
\textsuperscript{3}Key Laboratory of Computer Network and Information Integration (Southeast University), \\
~~Ministry of Education, China\\
}

\begin{document}

\maketitle

\begin{abstract}
Graph-based retrieval-augmented generation supports multi-hop retrieval by organizing corpus information into graphs. However, existing relation-free graph retrieval methods rely primarily on query–sentence similarity to search for evidence. This can exclude useful bridging evidence with low query similarity and activate incidental entities unrelated to the reasoning chain. In this paper, we propose a simple and effective approach called NexusRAG, which augments the relation-free Tri-Graph with a corpus-level entity neighborhood structure derived from joint entity co-occurrence and semantic similarity. NexusRAG employs this structure to guide two complementary propagation paths: neighborhood-constrained semantic propagation through sentences identifies the query-relevant entity frontier, while direct structural propagation between neighboring entities expands that frontier to structurally related entities. The propagated entity weights also inform neighborhood-aware passage initialization for Personalized PageRank. Experiments on three multi-hop QA benchmarks and a domain-specific subset of GraphRAG-Bench show that NexusRAG consistently outperforms existing approaches. On the GraphRAG-Bench subset, NexusRAG achieves the highest evidence recall in all question categories, exceeding baselines by 4.2--8.1 points.
\end{abstract}

\section{Introduction}
\label{sec:intro}

Retrieval-augmented generation (RAG) enables large language models (LLMs) to ground their responses in external evidence~\citep{lewis2020retrieval,gao2023retrieval}. Conventional RAG systems typically retrieve passages based on their semantic similarity to the query. Although effective in finding directly relevant information, this approach can struggle with questions that require connecting evidence across multiple documents. In such cases, necessary intermediate evidence may have a low similarity to the original query, with its relevance becoming apparent only through connections to other passages. Independent passage retrieval can therefore overlook useful bridging evidence, leaving the generator without sufficient information to answer the question~\citep{borgeaud2022improving,izacard2023atlas,han2024retrieval,zhang2025survey}.

Graph-based retrieval-augmented generation (GraphRAG) addresses this limitation by explicitly modeling connections within a corpus to guide evidence retrieval~\citep{edge2024local,procko2024graph,zhang2025survey}. Related approaches explore different forms of structural organization. RAPTOR~\citep{sarthi2024raptor} recursively clusters and summarizes text to build a hierarchical index, while Microsoft GraphRAG~\citep{edge2024local} constructs an entity graph and generates summaries of its communities. Other methods exploit explicit entity--relation graphs to support retrieval and reasoning~\citep{HippoRAG,gutiérrez2025hipporag2,guo2024lightrag,he2024g,luo2025gfm}. These structures help connect evidence across documents, allowing retrieval to account for relationships that independent passage scoring may overlook.

However, graph construction introduces its own sources of error. Inaccurate relation extraction can create misleading links between entities, while fragmented or inconsistent graph structures can hinder access to relevant evidence. Such errors can degrade the retrieved context and compromise downstream reasoning. GraphRAG-Bench~\citep{xiang2025use} evaluates graph construction, retrieval, and generation across four levels of task complexity: fact retrieval, complex reasoning, contextual summarization, and creative generation. In particular, its evidence recall metric measures the proportion of reference claims supported by the retrieved context, independently of the generated answer. This distinction highlights the importance of recovering complete supporting evidence when questions require integrating information across multiple passages.

These limitations motivate relation-free graph retrieval, which builds a useful corpus structure without explicitly extracting semantic relations between entities. LinearRAG~\citep{zhuang2025linearrag}, for example, constructs an entity-sentence-passage Tri-Graph using lightweight entity extraction and semantic linking. Starting from entities matched to the query, it iteratively propagates activation through the entity-sentence subgraph, guided by query-sentence similarity. The resulting entity scores, together with query-passage relevance, initialize Personalized PageRank (PPR)~\citep{haveliwala2002topic} on the entity-passage subgraph to rank passages. Corpus sentences thus serve as contextual bridges between entities, supporting multi-hop retrieval without explicit relation extraction.

However, LinearRAG’s sentence-mediated propagation ties entity activation to query-sentence similarity, which can lead to two failure modes. \emph{Query-gated cutoff} occurs when a sentence connecting relevant entities has low query similarity, weakening the activation signal and potentially preventing propagation from reaching necessary evidence. \emph{Spurious activation} occurs when a query-relevant sentence mentions incidental entities, allowing activation to spread to entities that do not support the answer. Together, these cases highlight a limitation of using sentence relevance to guide entity transitions: a useful bridge may have low query similarity, while a relevant sentence may contain irrelevant entities. This motivates complementing query-dependent sentence mediation with an explicit corpus-level entity neighborhood to guide propagation.

We propose NexusRAG, a corpus-guided graph retrieval framework that augments the relation-free Tri-Graph with a corpus-level neighbor prior derived from entity co-occurrence and semantic similarity. This prior captures entity relatedness without explicit relation extraction and guides a dual-path propagation process. The semantic propagation uses query-sentence similarity to identify a relevant entity frontier, with entity transitions constrained by the neighbor prior. The structural propagation then expands this frontier through structurally related entities. The resulting entity weights inform passage initialization for Personalized PageRank, incorporating neighborhood information into passage ranking. 

Our main contributions are as follows:

\begin{itemize}

\item We identify two failure modes of sentence-mediated entity propagation, \emph{query-gated cutoff} and \emph{spurious activation}, highlighting the limitations of query--sentence relevance as a guide for entity transitions.

\item We propose NexusRAG, a simple and effective approach which uses a corpus-level neighbor prior to guide dual-path propagation, recovering bridging evidence while suppressing spurious entity activation without explicit relation extraction.

\item Experiments on three multi-hop QA benchmarks and a domain-specific subset of GraphRAG-Bench demonstrate consistent improvements over multiple strong baselines. NexusRAG achieves the highest evidence recall in all four categories of the GraphRAG-Bench subset, exceeding baselines by 4.2--8.1 points.
\end{itemize}

\section{Related Work}
\label{sec:related}

\textbf{Retrieval-Augmented Generation (RAG)}  grounds language models on retrieved external evidence~\citep{lewis2020retrieval,karpukhin2020dense,gao2023retrieval}. Dense passage retrieval is effective when relevant information is localized, but questions that require evidence from multiple documents motivate retrieval methods that explicitly organize or reason over corpus structure.

\textbf{Graph-based Retrieval-Augmented Generation (GraphRAG)} introduces relational structure to support multi-hop retrieval. We focus on three graph construction strategies that are most relevant to this work.
%
One line of work organizes corpus information into hierarchical structures using clustering, community detection, or recursive summarization. Microsoft GraphRAG~\citep{edge2024local} identifies communities in an entity graph and generates summaries at multiple levels, while RAPTOR~\citep{sarthi2024raptor} recursively clusters and summarizes text to construct a hierarchical representation. E2GraphRAG~\citep{zhao20252graphrag} also uses hierarchical representations to support multi-hop retrieval. These methods provide access to corpus information at different levels of abstraction, but their performance depends on the intermediate grouping or summary structures.
%
Another line of work constructs explicit knowledge graphs by extracting entities and relations from text. HippoRAG~\citep{HippoRAG} and HippoRAG2~\citep{gutiérrez2025hipporag2} build knowledge graphs from extracted relations and apply Personalized PageRank for retrieval. LightRAG~\citep{guo2024lightrag} combines entity--relation structures with hierarchical representations, while G-Retriever~\citep{he2024g} and GFM-RAG~\citep{luo2025gfm} perform graph-based retrieval over structured textual knowledge. These approaches make relational information explicit, but relation extraction introduces additional indexing cost and can produce noisy or inconsistent relations.

\textbf{Relation-free Graph Retrieval} avoids explicit relation extraction and derive graph structure directly from corpus observations. LinearRAG~\citep{zhuang2025linearrag} constructs an entity--sentence--passage Tri-Graph using lightweight entity extraction and semantic linking. Its retrieval process activates entities through query-relevant sentences and aggregates passage importance with Personalized PageRank. NexusRAG retains this relation-free graph construction and introduces corpus-level neighbor weights computed from entity co-occurrence and semantic similarity. The same weights are used during entity propagation and passage initialization, providing a structural signal throughout retrieval.

\textbf{Reasoning-enhanced RAG} uses language models to decompose or refine complex queries rather than explicitly organizing the corpus into a graph. LogicRAG~\citep{logicrag} and LAG~\citep{lag} decompose questions into subqueries with logical dependencies, while Chain-of-Note~\citep{chainofNote} generates intermediate notes before retrieving supporting documents. Self-RAG~\citep{asai2024selfRAG} incorporates self-reflection into the retrieval-generation process. These approaches use query-side reasoning to retrieve evidence for multi-step questions, whereas NexusRAG uses corpus-side structural information for graph retrieval.

\begin{figure}[t]
 \centering
 \includegraphics[width=\linewidth]{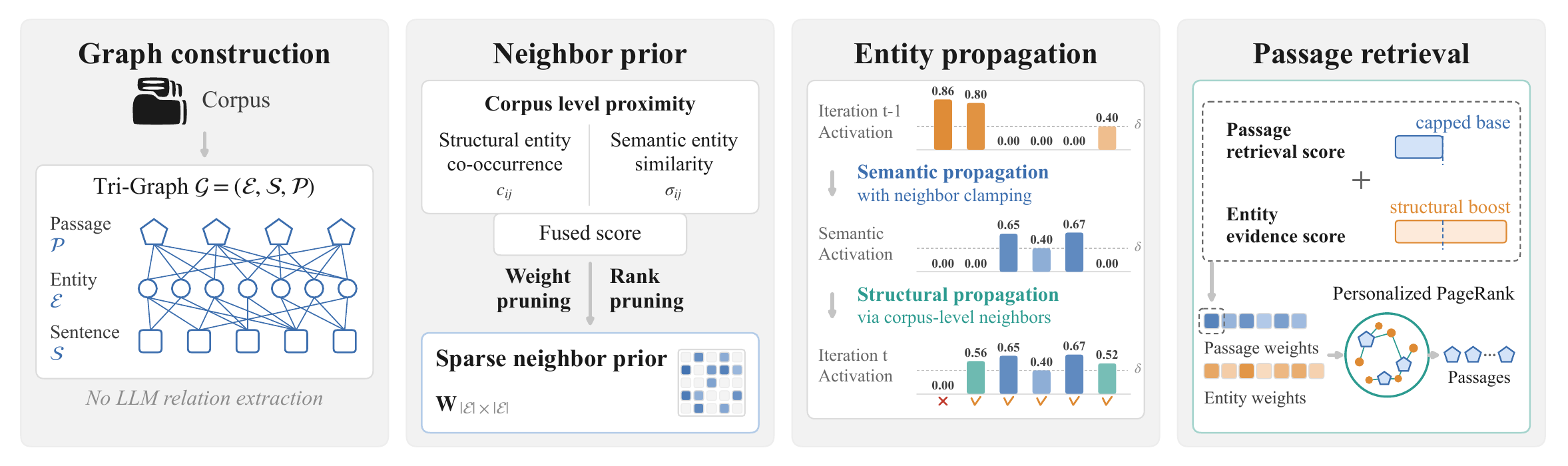}
\caption{Overview of NexusRAG. \textbf{1) Graph Construction.} The corpus is indexed into a relation-free entity--sentence--passage Tri-Graph. \textbf{2) Neighbor Prior.} A sparse corpus-level entity neighbor matrix $\matW$ is constructed from co-occurrence and semantic similarity with rank and weight pruning. \textbf{3) Entity Propagation.} Given a query $q$, neighbor-constrained semantic propagation identifies a query-relevant entity frontier, which structural propagation expands through $\matW$ to obtain the activated entity set. \textbf{4) Passage Retrieval.} Cumulative entity weights and neighbor-aware passage weights initialize Personalized PageRank for top-$K$ passage retrieval.}
 \label{fig:main_figure}
\vspace{-4mm}
\end{figure}

\section{Method}
\label{sec:method}

\textbf{Overview.} NexusRAG follows the relation-free GraphRAG paradigm established by LinearRAG~\citep{zhuang2025linearrag}, where entities serve as anchors for connecting evidence distributed across passages. The corpus is indexed as an entity--sentence--passage Tri-Graph $\gG$, with entity nodes $\gE$, sentence nodes $\gS$, and passage nodes $\gP$. Edges connect entities to the sentences and passages in which they appear, preserving contextual evidence within the original corpus structure. 

Retrieval proceeds in two stages: 1) entity activation, where query-related entities are activated and propagated through the entity--sentence subgraph to identify intermediate evidence, with each active entity selecting its top-$\eta$ query-similar sentences and propagating activation to other entities in those sentences with scores proportional to the source activation and query--sentence similarity; 2) passage retrieval, where the activated entity evidence is transferred to the entity--passage subgraph to initialize Personalized PageRank, which ranks the top-$K$ passages for the LLM.

This workflow places entity propagation at the center of multi-hop retrieval: the quality of the activated entities directly affects the evidence passed to passage initialization and subsequent PageRank ranking. We therefore revisit how query-dependent propagation determines entity importance and identify two failure modes: \textit{query-gated cutoff} and \textit{spurious activation}.

To address these limitations, we introduce NexusRAG with a query-independent neighbor prior that complements query-dependent sentence relevance with corpus-level entity structure. Figure~\ref{fig:main_figure} illustrates the overall framework. The neighbor prior supports a dual-path propagation process: the semantic propagation uses entity-level structural evidence to constrain transitions within query-relevant sentences, while the structural propagation expands activated entities through structurally related neighbors when sentence mediation is insufficient. The resulting entity evidence is then used for passage initialization and Personalized PageRank, providing a more reliable basis for multi-hop passage retrieval.

\subsection{Evidence Retrieval Failures in LinearRAG}
\label{sec:preliminary}

Multi-hop retrieval requires recovering intermediate evidence that connects successive reasoning steps. Results from GraphRAG-Bench~\citep{xiang2025use} reveal a less intuitive pattern: higher evidence recall does not necessarily lead to higher context relevance. With the passage retrieval budget fixed at $K=5$, one would expect retrieving more gold evidence to improve relevance if query relevance were sufficient to identify useful evidence. The observed mismatch suggests that evidence required for multi-hop reasoning is not always aligned with the query relevance used to guide retrieval.

\begin{figure}[t]
 \centering
 \includegraphics[width=\linewidth]{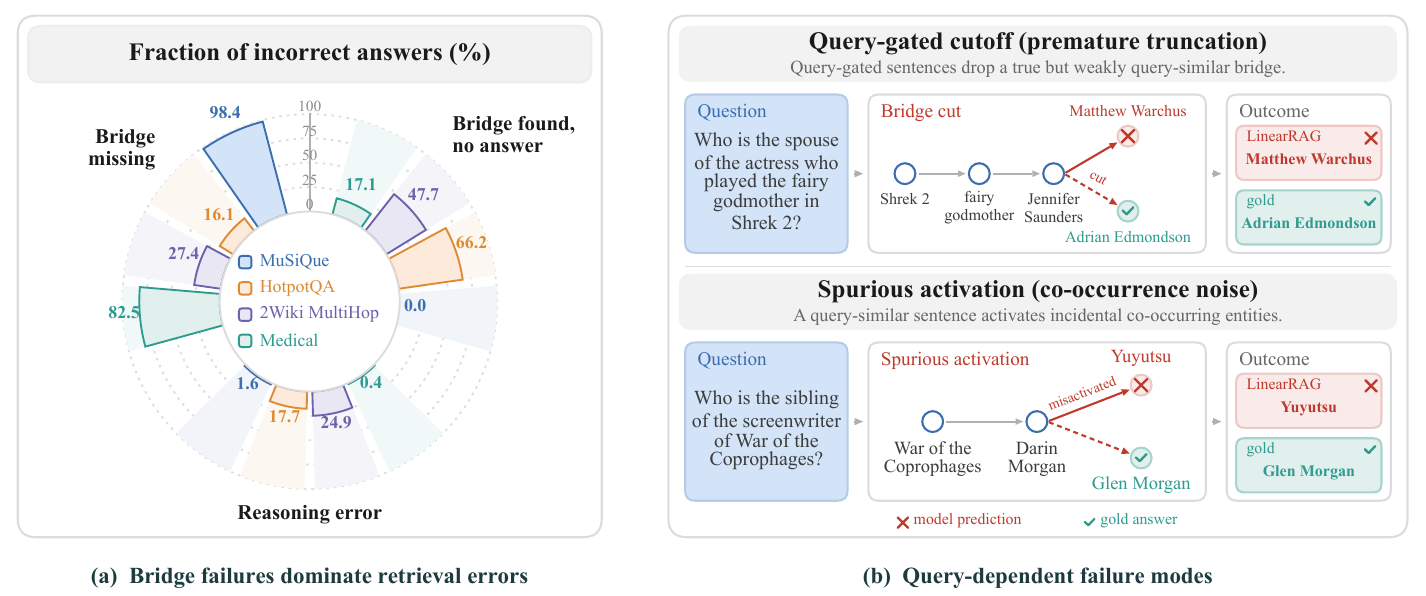}
 \caption{Retrieval failure analysis of LinearRAG. 
(a) Bridge entities account for a substantial fraction of retrieval errors. 
(b) Query-dependent sentence mediation can both suppress necessary bridge evidence and over-activate incidental entities.}
 \label{fig:pre}
\vspace{-4mm}
\end{figure}

We therefore examine LinearRAG, where entity propagation is mediated by query--sentence relevance. With GPT-4o-mini as the answer generator, LinearRAG achieves accuracies of $38.2\%$, $69.5\%$, $65.0\%$, and $65.32\%$ on MuSiQue, HotpotQA, 2WikiMultiHopQA, and the Medical subset of GraphRAG-Bench, respectively. Among its incorrect answers, $98.4\%$, $82.3\%$, $75.1\%$, and $99.6\%$ are classified as retrieval misses, respectively (Figure~\ref{fig:pre}(a)). This indicates that incomplete retrieval is a major source of failure and motivates examining how entity activation is propagated before passage ranking.

To locate the source of these retrieval misses, we further inspect the role of bridge evidence in the propagation process. For HotpotQA, 2WikiMultiHopQA, and Medical, we distinguish cases where the bridge entity is not retrieved from those where the bridge entity is retrieved but the answer passage is missing. The two cases account for $16.1\%$ and $66.2\%$ of incorrect answers on HotpotQA, $27.4\%$ and $47.7\%$ on 2WikiMultiHopQA, and $82.5\%$ and $17.1\%$ on Medical, respectively. These results show that retrieval failures are not limited to the final answer passage: in many cases, the intermediate bridge entity itself fails to receive sufficient activation for subsequent expansion.

We then examine LinearRAG's query-dependent sentence mediation (Figure~\ref{fig:pre}(b)) and identify two limitations.

\textbf{Query-gated cutoff.} LinearRAG uses query--sentence similarity to control sentence-mediated propagation. However, a sentence that contains necessary bridge evidence may be only indirectly related to the query and therefore receive a low similarity score. Its contribution can then fall below the pruning threshold before the bridge entity is sufficiently activated for further propagation. This indicates that query relevance at the sentence level does not necessarily reflect the importance of the evidence carried by that sentence.

\textbf{Spurious activation.} The opposite problem can also occur for highly query-relevant sentences. A single sentence may contain several entities, but these entities can play very different roles in the underlying evidence chain. Because sentence-mediated propagation derives their activation from the same query--sentence relevance, entities that are incidental to the reasoning chain can receive propagation mass together with the relevant entity. This indicates that sentence-level relevance alone cannot distinguish which entities within a relevant sentence should receive stronger propagation.

Together, these observations show that query--sentence relevance alone is insufficient to govern entity propagation: it can both suppress necessary bridge evidence and over-activate incidental entities. 

\subsection{Dual-Path Entity Propagation via Neighbor Prior}
\label{sec:dual_path}

To address these limitations, NexusRAG introduces a query-independent neighbor prior that captures corpus-level structural relatedness between entities. We incorporate this prior into two complementary propagation paths: the semantic propagation uses it to constrain entity transitions while retaining query relevance, and the structural propagation uses it to expand query-conditioned entities through structurally related neighbors.

\paragraph{Query-independent Neighbor Prior.}

We construct a query-independent neighbor prior from two complementary signals: sentence-level co-occurrence and semantic similarity between entity embeddings. Co-occurrence captures structural relatedness in the corpus: if two entities frequently appear in the same sentences, they are more likely to participate in the same local evidence or reasoning context. Semantic similarity links related entities in the embedding space, even when they rarely co-occur in the corpus. 

The co-occurrence signal is naturally sparse, with nonzero counts only for entity pairs that appear in the same sentence. To obtain a sparse semantic signal, we use approximate nearest-neighbor (ANN) search~\citep{malkov2020hnsw} to retain the top-$k$ most similar entities for each entity based on cosine similarity. We independently normalize the nonzero values of the two signals and combine them to obtain the neighbor weight between $e_i$ and $e_j$:
%
\begin{equation}
w_{ij}=\alpha\cdot c_{ij}+(1-\alpha)\cdot\sigma_{ij},
\label{eq:neighbor_fusion}
\end{equation}
where $c_{ij}$ and $\sigma_{ij}$ denote the normalized co-occurrence and cosine-similarity scores, respectively, and $\alpha\in[0,1]$ controls their relative contributions.

For each entity, we retain only neighbors whose fused scores exceed $\thresh$ and rank among its top-$\kappa$ candidates. All remaining entries, including self-connections, are set to zero. The resulting sparse neighbor matrix, $\matW=[w_{ij}]_{|\gE|\times|\gE|}$, is constructed offline and reused across queries.

At query time, we extract entities from $q$ using named entity recognition (NER) and match each extracted entity to its most similar entity in $\gE$. Matched entities are initialized with the corresponding similarity scores, while all other entities receive zero activation. NexusRAG then alternates between semantic and structural propagation. Semantic propagation uses the neighbor prior to guide updates through query-relevant sentence evidence, while structural propagation expands activations through corpus-level connections. The process terminates when all activation scores become zero or the maximum number of iterations is reached. We denote the final iteration by $T$.

\paragraph{Semantic propagation.}

Sentence-mediated propagation can cause \textit{spurious activation} by treating entities within a query-relevant sentence as equally relevant. NexusRAG uses the corpus-level neighbor prior to limit activation of weakly supported entities.
At iteration $t$, we initialize $\tilde{a}_j^{(t)}=0$ for every entity $e_j\in\gE$. For each entity $e_i$ activated in the previous iteration, we select the top-$\eta$ sentences containing it according to their cosine similarity to the query. A selected sentence with similarity $\sigma_m$ induces a candidate propagation score $a_i^{(t-1)}\sigma_m$. We discard candidates below the threshold $\delta$ and update the other entities in the sentence as follows:
\begin{equation}
\Tilde{a}^{(t)}_{j}=
\begin{cases}
\max\left(\min\left(a_{i}^{(t-1)}\cdot\sigma_{m},\,w_{ij}\right),\,\delta\right) & e_j\in\Neighbor(e_i),\\[4pt]
\delta & e_j\notin\Neighbor(e_i).
\end{cases}
\label{eq:hard_prop}
\end{equation}
Here, $w_{ij}$ and $\Neighbor(e_i)$ denote the neighbor weight and neighbor set defined by the corpus-level prior. Neighbor weights constrain propagation scores subject to a floor of $\delta$, while non-neighbors receive only this floor value. This attenuates incidental activation while preserving sentence-mediated connections. Updates are applied sequentially: when multiple admissible transitions reach the same entity, the latest assignment overwrites the previous score.

\paragraph{Structural propagation.}
Sentence-mediated propagation may stop when a sentence containing useful evidence has low query similarity, causing \textit{query-gated cutoff}. Structural propagation addresses this failure by expanding semantically activated entities through the corpus-level neighbor graph, allowing activation to reach additional entities.

We initialize $a_j^{(t)}=\tilde{a}_j^{(t)}$ and process neighbor transitions from entities activated during semantic propagation. A structural transition is accepted only if its target has not yet been activated in the current iteration and its score meets the threshold:
\begin{equation}
a^{(t)}_{j}=
\begin{cases}
w_{ij} \cdot \tilde{a}^{(t)}_{i}
& \tilde{a}^{(t)}_{j}=0
\;\land\;
w_{ij}\cdot \tilde{a}^{(t)}_{i}\geq\delta,\\
\tilde{a}^{(t)}_{j}
& \text{otherwise}.
\end{cases}
\label{eq:weak_prop}
\end{equation}
Thus, entities activated by semantic propagation retain their scores, while previously inactive entities receive the score of their first admissible structural transition. The resulting activations serve as inputs to the next iteration. We accumulate admitted activation scores across iterations to obtain $A_i$ for each entity $e_i\in\gE$.

\subsection{Neighbor-aware Passage Initialization}
\label{sec:hybrid_init}
A passage can be relevant to the query through either direct semantic similarity or entities reached during propagation. Because query--passage similarity alone can be misleading, we use a bounded dense retrieval score as a baseline and augment it with cumulative entity evidence from neighbor-aware propagation. The initial score of passage $p_j$ is
\begin{equation}
P_{\mathrm{init}}(p_j)=\exp({d}_j)+\sum_{e_i\in \gE}\frac{A_{i}\cdot\log\!\bigl(1+\mathrm{count}(e_i,p_j)\bigr)}{\max\!\bigl(\mathrm{hop}(e_i),1\bigr)},
\label{eq:hybrid_init}
\end{equation}
where $d_j\in[0,1]$ is the normalized query--passage cosine similarity, $A_i$ is the cumulative activation weight of entity $e_i$, $\mathrm{count}(e_i,p_j)$ counts its occurrences in passage $p_j$, and $\mathrm{hop}(e_i)$ denotes the propagation hop at which it was last activated.

The exponential transformation preserves the ordering of dense similarities. The entity term favors passages containing strongly activated entities, rewards repeated mentions with diminishing returns through logarithmic weighting, and discounts cumulative activation according to the last activation hop.
We use these passage scores together with the cumulative entity weights to initialize Personalized PageRank on the entity--passage subgraph. The top-$K$ passages in the resulting ranking are provided to the LLM for answering questions.

\section{Experiments}
\label{sec:experiments}
We organize our experiments around three research questions:
\textbf{Q1 (Effectiveness):} How does NexusRAG compare with the evaluated baselines in evidence recall and QA performance?
\textbf{Q2 (Component Analysis):} How do the corpus-level neighbor prior, dual-path propagation, and neighbor-aware passage initialization contribute to performance?
\textbf{Q3 (Sensitivity):} How do the fusion coefficient $\alpha$, pruning threshold $\thresh$, and neighbor cap $\kappa$ affect the calculation of neighbor prior?
Additional experiments on recall evaluation, efficiency and scalability, detailed sensitivity analyses, and case studies are provided in the appendix.

\begin{table}[t]
\centering
 \caption{Main results on four benchmarks across three generation backbones. All methods are evaluated with GPT-4o-mini; Qwen3.6-27B-FP8 and DeepSeek-V4-Flash provide additional comparisons between LinearRAG and NexusRAG. \textbf{Bold} and \underline{underline} denote the highest and second-highest results, respectively. Columns report Contain-Acc.\ (\%) (Con.), LLM-Acc.\ (\%) (LLM.), and Avg.\ (\%) (mean of Con.\ and LLM.). Only LLM-Acc.\ is used for the Medical dataset.}
 \label{tab:main_results}
 \setlength{\tabcolsep}{1mm}
 \scalebox{0.83}{
 \begin{tabular}{l*{10}{w{c}{1.15cm}}}
 \toprule
 \multirow{2}{*}{\textbf{Method}}
 &\multicolumn{3}{c}{\textbf{HotpotQA}}
 &\multicolumn{3}{c}{\textbf{2Wiki}}
 &\multicolumn{3}{c}{\textbf{MuSiQue}}
 &\textbf{Medical}\\
 \cmidrule(lr){2-4} \cmidrule(lr){5-7} \cmidrule(lr){8-10} \cmidrule(lr){11-11}
 &Con. &LLM. &Avg. &Con. &LLM. &Avg. &Con. &LLM. &Avg. &LLM. \\
 \midrule
 \multicolumn{11}{c}{\textbf{\textit{Direct Zero-shot LLM Inference}}} \\
 \midrule
 llama-8B & 31.10 & 27.30 & 29.20 & 33.60 & 16.20 & 24.90 & 7.40 & 8.10 & 7.75 & 27.31 \\
 llama-13B & 24.20 & 16.80 & 20.50 & 21.90 & 10.50 & 16.20 & 3.30 & 4.40 & 3.85 & 28.86 \\
 GPT-3.5-turbo & 33.40 & 43.20 & 38.30 & 28.70 & 31.00 & 29.85 & 10.30 & 21.90 & 16.10 & 45.60 \\
 GPT-4o-mini & 38.90 & 40.20 & 39.55 & 36.30 & 31.40 & 33.85 & 13.60 & 15.80 & 14.70 & 42.10 \\
 Qwen3.6-27B-FP8 & 37.80 & 38.70 & 38.25 & 50.20 & 34.30 & 42.25 & 15.50 & 15.10 & 15.30 & 40.60 \\
 \midrule
 \multicolumn{11}{c}{\textbf{\textit{Vanilla Retrieval-Augmented Generation}}} \\
 \midrule
 Retrieval (Top-1) & 46.30 & 49.10 & 47.70 & 36.60 & 31.70 & 34.15 & 17.80 & 21.10 & 19.45 & 48.01 \\
 Retrieval (Top-3) & 53.00 & 56.00 & 54.50 & 44.90 & 39.70 & 42.30 & 25.10 & 27.50 & 26.30 & 59.07 \\
 Retrieval (Top-5) & 55.70 & 58.60 & 57.15 & 48.60 & 43.00 & 45.80 & 26.10 & 29.60 & 27.85 & 61.68 \\
 \midrule
 \multicolumn{11}{c}{\textbf{\textit{Graph-based Retrieval-Augmented Generation Methods}}} \\
 \midrule
 KGP & 61.50 & 60.90 & 61.20 & 31.60 & 30.00 & 30.80 & 25.60 & 30.10 & 27.85 & 54.22\\
 G-Retriever & 42.20 & 40.60 & 41.40 & 46.60 & 27.10 & 36.85 & 14.40 & 15.50 & 14.95 & 50.36\\
 RAPTOR & 55.90 & 58.30 & 57.10 & 50.10 & 42.10 & 46.10 & 23.30 & 27.40 & 25.35 & 55.75\\
 E2GraphRAG & 61.00 & 63.90 & 62.45 & 54.30 & 38.10 & 46.20 & 23.80 & 26.20 & 25.00 & 58.00\\
 LightRAG & 60.30 & 59.50 & 59.90 & 55.20 & 39.00 & 47.10 & 27.40 & 28.60 & 28.00 & 54.36\\
 HippoRAG & 57.00 & 59.30 & 58.15 & 66.10 & 59.90 & 63.00 & 29.30 & 24.10 & 26.70 & 55.04\\
 GFM-RAG & 62.70 & 65.60 & 64.15 & 66.80 & 59.60 & 63.20 & 29.90 & 34.60 & 32.25 & 56.07\\
 HippoRAG2 & 62.90 & 64.30 & 63.60 & 62.70 & 55.00 & 58.85 & 31.00 & 35.00 & 33.00 & 60.77\\
 \midrule
 \multicolumn{11}{c}{\textbf{\textit{Linear Graph Retrieval-Augmented Generation Methods}}} \\
 \midrule
 LinearRAG & \underline{65.5} & \underline{69.5} & \underline{67.50} & \underline{69.5} & \underline{65.0} & \underline{67.25} & \underline{32.3} & \underline{38.2} & \underline{35.25} & \underline{65.32}\\

 \shortstack[l]{LinearRAG\\+ DeepSeek-V4-Flash} & \underline{76.0} & \underline{86.7} & \underline{81.35} & \underline{83.3} & \underline{86.8} & \underline{85.05} & \underline{46.7} & \underline{57.5} & \underline{52.10} & \underline{66.00}\\

 \shortstack[l]{LinearRAG\\+ Qwen3.6-27B-FP8} & \underline{69.3} & \underline{85.0} & \underline{77.15} & \underline{80.0} & \underline{83.1} & \underline{81.55} & \underline{45.6} & \underline{54.5} & \underline{50.05} & \underline{70.22}\\
 \midrule
 {NexusRAG (ours)} & \textbf{70.1} & \textbf{72.9} & \textbf{71.50} & \textbf{72.2} & \textbf{68.7} & \textbf{70.45} & \textbf{35.5} & \textbf{41.0} & \textbf{38.25} & \textbf{73.91}\\
{\shortstack[l]{NexusRAG (ours)\\+ DeepSeek-V4-Flash}} & \textbf{79.8} & \textbf{88.1} & \textbf{83.95} & \textbf{85.9} & \textbf{87.6} & \textbf{86.75} & \textbf{50.0} & \textbf{60.4} & \textbf{55.20} & \textbf{78.90}\\

 {\shortstack[l]{NexusRAG (ours)\\+ Qwen3.6-27B-FP8}} & \textbf{72.2} & \textbf{88.7} & \textbf{80.45} & \textbf{81.4} & \textbf{86.0} & \textbf{83.70} & \textbf{47.1} & \textbf{58.4} & \textbf{52.75} & \textbf{73.96}\\
 \bottomrule
 \end{tabular}}
\end{table}

\subsection{Experimental Setting}

\textbf{Datasets.} We evaluate NexusRAG on three multi-hop QA benchmarks (HotpotQA, 2WikiMultiHopQA, and MuSiQue) and the Medical subset of GraphRAG-Bench~\citep{xiang2025use}. For the three multi-hop benchmarks, we adopt the 1,000-question validation subsets used by LinearRAG and HippoRAG~\citep{zhuang2025linearrag,HippoRAG}. For Medical, we evaluate 2,062 questions following the GraphRAG-Bench evaluation protocol.

\textbf{Baselines.} We compare NexusRAG with Vanilla RAG (Top-1, Top-3, and Top-5) and the following graph-based retrieval methods: KGP~\citep{wang2024knowledge}, G-Retriever~\citep{he2024g}, RAPTOR~\citep{sarthi2024raptor}, E$^{2}$GraphRAG~\citep{zhao20252graphrag}, LightRAG~\citep{guo2024lightrag}, HippoRAG~\citep{HippoRAG}, GFM-RAG~\citep{luo2025gfm}, and HippoRAG2~\citep{gutiérrez2025hipporag2}. LinearRAG~\citep{zhuang2025linearrag} serves as our direct baseline. Details of all baselines are provided in Appendix~\ref{app:baselines}.

\textbf{Metrics.} Following \citet{zhuang2025linearrag}, we evaluate QA performance using two metrics. \emph{Contain-Match Accuracy (Contain-Acc.)} measures the proportion of generated responses containing the reference answer. \emph{LLM-as-a-Judge Accuracy (LLM-Acc.)} evaluates answer correctness against the reference, allowing for valid paraphrases and formatting differences. We report both metrics on the three multi-hop QA benchmarks and only LLM-Acc. on Medical, whose reference answers are typically long and descriptive. For retrieval quality, we adopt \emph{Context Relevance} and \emph{Evidence Recall} from GraphRAG-Bench~\citep{xiang2025use}. Context Relevance assesses the alignment between the retrieved context and the question, while Evidence Recall measures the proportion of reference claims supported by the retrieved context. Retrieval results are reported at the end of the Q1 evaluation.

\textbf{Implementation.} We use all-mpnet-base-v2 \citep{song2020mpnet} for extracting entity, sentence, and passage embeddings, with spaCy \citep{honnibal2020spacy} en\_core\_web\_trf for entity recognition on the general-domain datasets and  en\_core\_sci\_scibert for the Medical corpus. We retrieve $K=5$ passages by default. Table~\ref{tab:main_results} reports results with GPT-4o-mini, DeepSeek-V4-Flash, and Qwen3.6-27B-FP8, with each model serving as both generator and judge in its corresponding setting. For the neighbor prior, we set $\alpha=0.5$, $\thresh=0.5$, and $\kappa=5$, with the ANN candidate count set to $k=\kappa$. Retaining at most $\kappa$ neighbors per entity limits index storage and the computation required for direct neighbor expansion. The propagation and PPR parameters follow LinearRAG's reported configuration. Parameter sensitivity is examined in Section~\ref{sec:param_sensitivity}.

\subsection{Main Results and Evidence Retrieval (Q1)}

Table~\ref{tab:main_results} presents the main comparison. Relative to LinearRAG, NexusRAG obtains higher Avg. scores on all four datasets, while the additional backbones show the same direction of change.

%
\obs The main results show that several graph-based retrieval already provides an advantage over conventional RAG, while the relation-free LinearRAG further demonstrates that explicit relation extraction is not necessary to obtain effective graph-based retrieval. Nevertheless, LinearRAG still leaves room for improvement. On GPT-4o-mini, NexusRAG consistently improves the overall Avg.\ over LinearRAG on all three datasets with both metrics, by +4.0\% on HotpotQA, +3.0\% on MuSiQue, and +3.2\% on 2Wiki. Both Contain-Acc.\ and LLM-Acc.\ increase on these datasets; for example, HotpotQA improves by +4.6\% and +3.4\%, respectively. On Medical, where only LLM-Acc.\ is reported, NexusRAG further improves it by 8.6\%. These results indicate that although relation-free graph retrieval already alleviates limitations of conventional RAG, the entity propagation used by LinearRAG remains an important source of retrieval error.

\obs The same trend holds with Qwen3.6-27B-FP8, suggesting that the improvement is not tied to a particular generation backbone. NexusRAG improves Avg.\ over LinearRAG by +3.30\% on HotpotQA, +2.15\% on 2Wiki, and +2.70\% on MuSiQue, while LLM-Acc.\ on Medical increases by +3.74\%. The consistency across datasets and backbones suggests that NexusRAG improves the retrieval process itself rather than merely exploiting differences in the generator.

To evaluate retrieval quality independently of generation, we follow the GraphRAG-Bench protocol on the Medical dataset, which contains four question categories of increasing difficulty: Fact Retrieval, Complex Reasoning, Contextual, and Creative Generation. The results are reported in Table~\ref{tab:retrieval_results}.

\begin{table*}[t]
 \caption{{Retrieval quality evaluation results (\%) following the GraphRAG-Bench protocol across four question categories.} Recall and relevance are reported for fact retrieval, complex reasoning, contextual understanding, and creative generation. The highest and second-highest results are shown in \textbf{bold} and \underline{underlined}, respectively.}
 \label{tab:retrieval_results}
 \small
 \centering
 \setlength{\tabcolsep}{1mm}
 \begin{tabular}{lcccccccc}
 \toprule
 \multirow{2}{*}{\textbf{Method}}
 &\multicolumn{2}{c}{\textbf{Fact Retrieval}}
 & \multicolumn{2}{c}{\textbf{Complex Reasoning}}
 &\multicolumn{2}{c}{\textbf{Contextual}}
 & \multicolumn{2}{c}{\textbf{Creative Generation}}\\
 \cmidrule(lr){2-3} \cmidrule(lr){4-5} \cmidrule(lr){6-7} \cmidrule(lr){8-9}
 &Recall &Relevance &Recall &Relevance &Recall &Relevance &Recall &Relevance \\
 \midrule
 Vanilla RAG (Top-5) & 86.24 & 63.71 & 84.97 & \textbf{84.11} & 84.14 & \textbf{89.94} & 44.88 & 58.73 \\
 RAPTOR & 85.40 & 69.38 & \underline{89.70} & 53.20 & 88.86 & 58.73 & 72.70 & 52.71 \\
 E$^2$GraphRAG & 87.84 & 69.74 & 87.08 & 62.67 & \underline{89.17} & 71.63 & 60.26 & 35.84 \\
 LightRAG & 80.32 & 41.27 & 82.91 & 42.79 & 85.71 & 43.11 & 81.34 & 45.17 \\
 GFM-RAG & \underline{90.08} & 57.90 & 85.03 & 33.06 & 78.62 & 40.14 & 83.51 & 22.87 \\
 HippoRAG & 87.25 & 52.44 & 83.80 & 42.19 & 83.46 & 49.13 & 81.66 & 45.03 \\
 LinearRAG & 88.86 & \textbf{86.09} & 87.03 & \underline{81.58} & 89.13 & \underline{87.89} & \underline{89.08} & \textbf{72.74} \\
 \midrule
 {NexusRAG (ours)} & \textbf{94.30} & \underline{81.17} & \textbf{94.62} & 68.52 & \textbf{94.36} & 79.41 & \textbf{97.16} & \underline{61.66} \\
 \bottomrule
 \end{tabular}
 \vspace{-2mm}
\end{table*}

\obs Table~\ref{tab:retrieval_results} reports two retrieval metrics for the baselines, LinearRAG, and NexusRAG: \emph{recall} (evidence recall, the fraction of all gold evidence retrieved) and \emph{relevance} (context relevancy, the relevance of the retrieved content to the query). NexusRAG obtains the highest evidence recall in all four categories, with differences of 4.2--8.1 points relative to the next-highest result in each category. The context-relevance scores vary across question categories, from 61.66 on Creative Generation to 81.17 on Fact Retrieval. This variation is consistent with the retrieval objective: recovering bridge evidence can add content that is necessary for the evidence chain but less directly aligned with the query wording.

\begin{wraptable}{r}{0.46\linewidth}
\vspace{-4mm}
\centering
\caption{{LLM-Acc.\ (\%) of the same GPT-4o-mini predictions re-judged by three evaluators.} Highest results are in \textbf{bold}.}
\label{tab:evaluator_robust}
\setlength{\tabcolsep}{1mm}
\resizebox{\linewidth}{!}{
\begin{tabular}{llcccc}
\toprule
\textbf{Method} & \textbf{Evaluator} & \textbf{Hotpot} & \textbf{2Wiki} & \textbf{MuSiQue} & \textbf{Med.} \\
\midrule
\multirow{3}{*}{LinearRAG}
 & GPT-4o-mini & 69.5 & 65.0 & 38.2 & 65.3 \\
 & DeepSeek-V4-Flash & 78.1 & 73.8 & 42.8 & 60.5 \\
 & Qwen3.6-27B-FP8 & 78.4 & 74.9 & 42.9 & 74.3 \\
\midrule
\multirow{3}{*}{NexusRAG}
 & GPT-4o-mini & \textbf{72.9} & \textbf{68.7} & \textbf{41.0} & \textbf{73.91} \\
 & DeepSeek-V4-Flash & \textbf{80.2} & \textbf{74.9} & \textbf{44.6} & \textbf{63.06} \\
 & Qwen3.6-27B-FP8 & \textbf{81.9} & \textbf{76.7} & \textbf{46.6} & \textbf{79.29} \\
\bottomrule
\end{tabular}}
\vspace{-4mm}
\end{wraptable}

\obs (evaluator comparison). Table~\ref{tab:main_results} shows higher LLM-Acc.\ when the generator and evaluator are the same model (e.g., $88.7$ on HotpotQA for NexusRAG with Qwen3.6-27B-FP8). To separate generation from evaluation, we fix the GPT-4o-mini predictions and re-judge the same outputs with three evaluators: GPT-4o-mini, DeepSeek-V4-Flash, and Qwen3.6-27B-FP8. Table~\ref{tab:evaluator_robust} reports the resulting LLM-Acc.\ scores. NexusRAG scores higher than LinearRAG for each evaluator on each dataset. Absolute scores differ across evaluators, reflecting differences in evaluator calibration.

\subsection{Ablation Study (Q2)}
\label{sec:ablation}

The neighbor prior $\matW$ is consulted at three points of the pipeline, and we ablate each in turn. \textbf{(1)}~w/o Neighbor removes $\matW$ altogether. \textbf{(2)}~w/o Structural propagation drops its \emph{expansion} use. \textbf{(3)}~w/o Neighbor Clamp drops its \emph{gating} use, so non-neighbor transitions are no longer held at the $\delta$ floor. \textbf{(4)}~w/o Neighbor-aware Init drops its readout, initializing passages without the neighbor-aware entity weights of Eq.~\ref{eq:hybrid_init}. Results are in Table~\ref{tab:ablation}.

\begin{wraptable}{r}{0.48\linewidth}
\centering
\vspace{-4mm}
\caption{{Ablation across the four benchmarks (Qwen3.6-27B-FP8).}}
\label{tab:ablation}
\setlength{\tabcolsep}{1mm}
\resizebox{\linewidth}{!}{%
\begin{tabular}{lccccccc}
\toprule
& \multicolumn{2}{c}{\textbf{HotpotQA}} & \multicolumn{2}{c}{\textbf{2Wiki}} & \multicolumn{2}{c}{\textbf{MuSiQue}} & \textbf{Med.} \\
\cmidrule(lr){2-3} \cmidrule(lr){4-5} \cmidrule(lr){6-7} \cmidrule(lr){8-8}
\textbf{Variant} & \textbf{Con.} & \textbf{LLM.} & \textbf{Con.} & \textbf{LLM.} & \textbf{Con.} & \textbf{LLM.} & \textbf{LLM.} \\
\midrule
NexusRAG & \textbf{72.2} & \textbf{88.7} & \textbf{81.4} & \textbf{86.0} & \textbf{47.1} & \textbf{58.4} & \textbf{73.96} \\
w/o Neighbor Clamp & 71.2 & 86.6 & 80.0 & 83.4 & 46.2 & 56.9 & 72.58 \\
w/o Structural propagation & 71.4 & 87.7 & 80.8 & 84.8 & 45.0 & 55.0 & 71.10 \\
w/o Neighbor & 69.8 & 85.6 & 79.8 & 82.8 & 45.7 & 54.6 & 70.81 \\
w/o Neighbor-aware Init & 71.3 & 87.6 & 80.9 & 84.9 & 46.1 & 56.3 & 72.50 \\
\bottomrule
\end{tabular}}
\vspace{-4mm}
\end{wraptable}
\obs Each use of $\matW$ contributes to the final score. Removing Neighbor Clamp, Structural propagation, or Neighbor-aware Init changes average LLM-Acc.\ by $-1.9$, $-2.1$, and $-1.4$ points, respectively, while removing $\matW$ altogether changes it by $-3.3$ points. The three uses therefore provide complementary contributions, with the full prior giving the largest change among the ablations.

\subsection{Parameter Sensitivity (Q3)}
\label{sec:param_sensitivity}
\begin{wrapfigure}{r}{0.50\linewidth}
\centering
\vspace{-6mm}
\includegraphics[width=\linewidth]{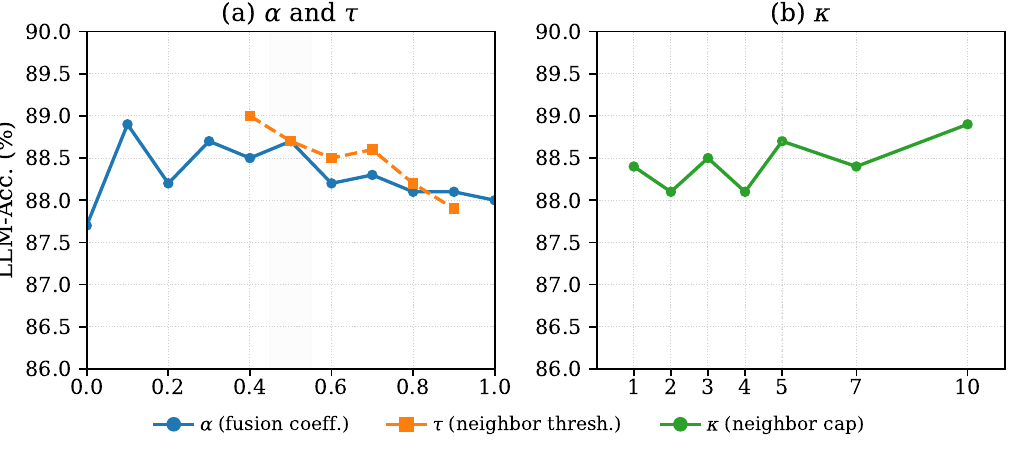}
\caption{Sensitivity study of the neighbor-mechanism parameters on the HotpotQA dataset.}
\label{fig:sensitivity_llm}
\vspace{-3mm}
\end{wrapfigure}

We vary $\alpha$, $\thresh$, and $\kappa$ one at a time on HotpotQA with the Qwen3.6-27B-FP8 backbone (Figure~\ref{fig:sensitivity_llm}); full details are in Appendix~\ref{app:param_detail}.

\obs Across the swept $\alpha$ range, LLM-Acc.\ varies by at most $1.2$ points; within $[0.1,0.9]$ it remains in the range $88.1$--$88.9$, while $\alpha=0$ and $\alpha=1$ give $87.7$ and $88.0$, respectively. Across the swept $\thresh$ range, LLM-Acc.\ changes from $89.0$ at $\thresh=0.4$ to $87.9$ at $\thresh=0.9$. The density drops by about $66\times$ between $\thresh=0.5$ and $\thresh=0.6$, while LLM-Acc.\ changes by $0.2$ points. $\kappa$ varies within $88.1$--$88.9$ in the tested range.

\section{Conclusion}
\label{sec:conclusion}

We propose NexusRAG, a relation-free graph retrieval framework guided by a corpus-level neighbor prior derived from entity co-occurrence and semantic similarity. The prior guides dual-path entity propagation and passage initialization for Personalized PageRank to mitigate query-gated cutoff and spurious activation. Experiments on three multi-hop QA benchmarks and a domain-specific subset of GraphRAG-Bench show that NexusRAG consistently outperforms the evaluated baselines across retrieval paradigms. On the GraphRAG-Bench subset, NexusRAG achieves the highest evidence recall in all four question categories, exceeding the next-best result in each category by 4.2--8.1 points.
\newpage
\section*{AI use statement}

Large language models (LLMs) were used to aid in writing and polishing the manuscript, including sentence rephrasing, grammar checking, and improving readability and flow. The LLM was not involved in ideation, research methodology, or experimental design; all concepts, analyses, and results are developed and conducted by the authors. We take full responsibility for the entire manuscript, including any text refined with LLM assistance, and have ensured it adheres to ethical guidelines without plagiarism or scientific misconduct.

\section*{Ethics Statement}
The four datasets used in the experiments: HotpotQA, 2WikiMultiHopQA, MuSiQue and Medical, are widely used public benchmarks. Our research strictly adheres to the ICLR Code of Ethics, particularly regarding data privacy, transparency, and responsible computing practices. No human participants were involved in this study.

\section*{Reproducibility Statement}
We provide the full experimental configuration in Table~\ref{tab:hyperparams} (Appendix), and the complete neighbor construction, dual-path propagation, and passage initialization procedures are specified in Eq.~\ref{eq:neighbor_fusion}--\ref{eq:hybrid_init}. The implementation follows the experimental settings described in Section~\ref{sec:experiments}. Code will be released upon publication.

\bibliography{iclr2027_conference}

@string{emnlp = "Empirical Methods in Natural Language Processing (EMNLP)"}

@string{neurips = "Advances in Neural Information Processing Systems (NeurIPS)"}

@string{icml = "International Conference on Machine Learning (ICML)"}

@string{aaai = "Conference on Artificial Intelligence (AAAI)"}

@string{iclr = "International Conference on Learning Representations (ICLR)"}

@string{www = "International World Wide Web Conference (WWW)"}

@string{jmlr = "The Journal of Machine Learning Research (JMLR)"}

@inproceedings{2wikimqa,
  title={Constructing {A} Multi-hop {QA} Dataset for Comprehensive Evaluation of Reasoning Steps},
  author={Ho, Xanh and Duong Nguyen, Anh-Khoa and Sugawara, Saku and Aizawa, Akiko},
  booktitle={Proceedings of the 28th International Conference on Computational Linguistics},
  pages={6609--6625},
  address={Barcelona, Spain (Online)},
  publisher={International Committee on Computational Linguistics},
  month=dec,
  year={2020}
}

@article{trivedi2022musique,
  title={MuSiQue: Multi-hop Questions via Single-hop Question Composition},
  author={Trivedi, Harsh and Balasubramanian, Niranjan and Khot, Tushar and Sabharwal, Ashish},
  journal={Transactions of the Association for Computational Linguistics},
  volume={10},
  pages={539--554},
  year={2022}
}

@inproceedings{wang2024knowledge,
  title={Knowledge graph prompting for multi-document question answering},
  author={Wang, Yu and Lipka, Nedim and Rossi, Ryan A and Siu, Alexa and Zhang, Ruiyi and Derr, Tyler},
  booktitle=aaai,
  year={2024}
}

@article{edge2024local,
  title={From local to global: A graph rag approach to query-focused summarization},
  author={Edge, Darren and Trinh, Ha and Cheng, Newman and Bradley, Joshua and Chao, Alex and Mody, Apurva and Truitt, Steven and Larson, Jonathan},
  journal={arXiv preprint arXiv:2404.16130},
  year={2024}
}

@inproceedings{he2024g,
  title={G-retriever: Retrieval-augmented generation for textual graph understanding and question answering},
  author={He, Xiaoxin and Tian, Yijun and Sun, Yifei and Chawla, Nitesh V and Laurent, Thomas and LeCun, Yann and Bresson, Xavier and Hooi, Bryan},
  booktitle={Advances in Neural Information Processing Systems (NeurIPS)},
  year={2024}
}

@article{gao2023retrieval,
  title={Retrieval-augmented generation for large language models: A survey},
  author={Gao, Yunfan and Xiong, Yun and Gao, Xinyu and Jia, Kangxiang and Pan, Jinliu and Bi, Yuxi and Dai, Yi and Sun, Jiawei and Wang, Haofen},
  journal={arXiv preprint arXiv:2312.10997},
  year={2023}
}

@inproceedings{borgeaud2022improving,
  title={Improving language models by retrieving from trillions of tokens},
  author={Borgeaud, Sebastian and Mensch, Arthur and Hoffmann, Jordan and Cai, Trevor and Rutherford, Eliza and Millican, Katie and Van Den Driessche, George Bm and Lespiau, Jean-Baptiste and Damoc, Bogdan and Clark, Aidan and others},
  booktitle=icml,
  year={2022}
}

@inproceedings{karpukhin2020dense,
  title={Dense passage retrieval for open-domain question answering},
  author={Karpukhin, Vladimir and O{\u{g}}uz, Barlas and Min, Sewon and Lewis, Patrick and Wu, Ledell and Edunov, Sergey and Chen, Danqi and Yih, Wen-tau},
  booktitle={emnlp},
  pages={6769--6781},
  year={2020}
}

@article{izacard2023atlas,
  title={Atlas: Few-shot learning with retrieval augmented language models},
  author={Izacard, Gautier and Lewis, Patrick and Lomeli, Maria and Hosseini, Lucas and Petroni, Fabio and Schick, Timo and Dwivedi-Yu, Jane and Joulin, Armand and Riedel, Sebastian and Grave, Edouard},
  journal=jmlr,
  year={2023}
}

@inproceedings{luo2025gfm,
  title={GFM-RAG: Graph Foundation Model for Retrieval Augmented Generation},
  author={Luo, Linhao and Zhao, Zicheng and Haffari, Gholamreza and Phung, Dinh and Gong, Chen and Pan, Shirui},
  booktitle={Advances in Neural Information Processing Systems (NeurIPS)},
  year={2025}
}

@article{zhao20252graphrag,
  title={E2GraphRAG: Streamlining Graph-based RAG for High Efficiency and Effectiveness},
  author={Zhao, Yibo and Zhu, Jiapeng and Guo, Ye and He, Kangkang and Li, Xiang},
  journal={arXiv preprint arXiv:2505.24226},
  year={2025}
}

@inproceedings{guo2024lightrag,
  title={{L}ight{RAG}: Simple and Fast Retrieval-Augmented Generation},
  author={Guo, Zirui and Xia, Lianghao and Yu, Yanhua and Ao, Tu and Huang, Chao},
  booktitle={Findings of the Association for Computational Linguistics: EMNLP 2025},
  pages={10746--10761},
  address={Suzhou, China},
  publisher={Association for Computational Linguistics},
  month=nov,
  year={2025},
  doi={10.18653/v1/2025.findings-emnlp.568}
}

@inproceedings{asai2024selfRAG,
  title={Self-{RAG}: Learning to Retrieve, Generate, and Critique through Self-Reflection},
  author={Asai, Akari and Wu, Zeqiu and Wang, Yizhong and Sil, Avirup and Hajishirzi, Hannaneh},
  booktitle=iclr,
  pages={9112--9141},
  year={2024}
}

@inproceedings{chainofNote,
    title = "Chain-of-Note: Enhancing Robustness in Retrieval-Augmented Language Models",
    author = "Yu, Wenhao  and
      Zhang, Hongming  and
      Pan, Xiaoman  and
      Cao, Peixin  and
      Ma, Kaixin  and
      Li, Jian  and
      Wang, Hongwei  and
      Yu, Dong",
    editor = "Al-Onaizan, Yaser  and
      Bansal, Mohit  and
      Chen, Yun-Nung",
    booktitle = "Proceedings of the 2024 Conference on Empirical Methods in Natural Language Processing",
    month = nov,
    year = "2024",
    pages = "14672--14685",
}

@inproceedings{procko2024graph,
  title={Graph Retrieval-Augmented Generation for Large Language Models: A Survey},
  author={Procko, Tyler Thomas and Ochoa, Omar},
  booktitle={2024 Conference on AI, Science, Engineering, and Technology (AIxSET)},
  pages={166--169},
  publisher={IEEE},
  month=sep,
  year={2024},
  doi={10.1109/AIxSET62544.2024.00030}
}

@article{zhang2025survey,
  title={A Survey of Graph Retrieval-Augmented Generation for Customized Large Language Models},
  author={Zhang, Qinggang and Chen, Shengyuan and Bei, Yuanchen and Yuan, Zheng and Zhou, Huachi and Hong, Zijin and Dong, Junnan and Chen, Hao and Chang, Yi and Huang, Xiao},
  journal={arXiv preprint arXiv:2501.13958},
  year={2025}
}

@article{han2024retrieval,
  title={Retrieval-augmented generation with graphs (graphrag)},
  author={Han, Haoyu and Wang, Yu and Shomer, Harry and Guo, Kai and Ding, Jiayuan and Lei, Yongjia and Halappanavar, Mahantesh and Rossi, Ryan A and Mukherjee, Subhabrata and Tang, Xianfeng and others},
  journal={arXiv preprint arXiv:2501.00309},
  year={2025}
}

@inproceedings{yang2018hotpotqa,
  title={HotpotQA: A dataset for diverse, explainable multi-hop question answering},
  author={Yang, Zhilin and Qi, Peng and Zhang, Saizheng and Bengio, Yoshua and Cohen, William W and Salakhutdinov, Ruslan and Manning, Christopher D},
  booktitle=emnlp,
  year={2018}
}

@inproceedings{logicrag,
  title={You Don't Need Pre-built Graphs for {RAG}: Retrieval Augmented Generation with Adaptive Reasoning Structures},
  author={Chen, Shengyuan and Zhou, Chuang and Yuan, Zheng and Zhang, Qinggang and Cui, Zeyang and Chen, Hao and Xiao, Yilin and Cao, Jiannong and Huang, Xiao},
  booktitle=aaai,
  year={2026}
}

@article{lag,
  title={{LAG}: Logic-Augmented Generation from a Cartesian Perspective},
  author={Xiao, Yilin and Zhou, Chuang and Zhang, Qinggang and Dong, Su and Chen, Shengyuan and Huang, Xiao},
  journal={arXiv preprint arXiv:2508.05509},
  year={2025}
}

@inproceedings{HippoRAG,
     author = {Guti\'{e}rrez, Bernal Jim\'{e}nez and Shu, Yiheng and Gu, Yu and Yasunaga, Michihiro and Su, Yu},
     booktitle = neurips,
     title = {HippoRAG: Neurobiologically Inspired Long-Term Memory for Large Language Models},
     year = {2024}
}

@inproceedings{gutiérrez2025hipporag2,
  title={From rag to memory: Non-parametric continual learning for large language models},
  author={Guti{\'e}rrez, Bernal Jim{\'e}nez and Shu, Yiheng and Qi, Weijian and Zhou, Sizhe and Su, Yu},
  booktitle={International Conference on Machine Learning (ICML)},
  year={2025}
}

@inproceedings{sarthi2024raptor,
    title={RAPTOR: Recursive Abstractive Processing for Tree-Organized Retrieval},
    author={Sarthi, Parth and Abdullah, Salman and Tuli, Aditi and Khanna, Shubh and Goldie, Anna and Manning, Christopher D.},
    booktitle={International Conference on Learning Representations (ICLR)},
    year={2024}
}

@inproceedings{lewis2020retrieval,
  title={Retrieval-augmented generation for knowledge-intensive nlp tasks},
  author={Lewis, Patrick and Perez, Ethan and Piktus, Aleksandra and Petroni, Fabio and Karpukhin, Vladimir and Goyal, Naman and K{\"u}ttler, Heinrich and Lewis, Mike and Yih, Wen-tau and Rockt{\"a}schel, Tim and others},
  booktitle=neurips,
  year={2020}
}

@inproceedings{xiang2025use,
  title={When to Use Graphs in {RAG}: A Comprehensive Analysis for Graph Retrieval-Augmented Generation},
  author={Xiang, Zhishang and Wu, Chuanjie and Zhang, Qinggang and Chen, Shengyuan and Hong, Zijin and Huang, Xiao and Su, Jinsong},
  booktitle=iclr,
  year={2026}
}

@inproceedings{zhuang2025linearrag,
  title={LinearRAG: Linear Graph Retrieval Augmented Generation on Large-scale Corpora},
  author={Zhuang, Luyao and Chen, Shengyuan and Xiao, Yilin and Zhou, Huachi and Zhang, Yujing and Chen, Hao and Zhang, Qinggang and Huang, Xiao},
  booktitle={International Conference on Learning Representations (ICLR)},
  year={2026}
}

@inproceedings{haveliwala2002topic,
  title={Topic-Sensitive {PageRank}},
  author={Haveliwala, Taher H.},
  editor={Lassner, D. and Roure, D. D. and Iyengar, A.},
  booktitle={Proceedings of the Eleventh International World Wide Web Conference (WWW 2002)},
  address={Honolulu, Hawaii, USA},
  pages={517--526},
  year={2002},
  publisher={ACM},
  doi={10.1145/511446.511513},
  url={https://dl.acm.org/doi/10.1145/511446.511513}
}

@article{song2020mpnet,
title={Mpnet: Masked and permuted pre-training for language understanding},
author={Song, Kaitao and Tan, Xu and Qin, Tao and Lu, Jianfeng and Liu, Tie-Yan},
journal={Advances in neural information processing systems},
volume={33},
pages={16857--16867},
year={2020}
}

@misc{honnibal2020spacy,
  title={{spaCy}: Industrial-strength Natural Language Processing in {Python}},
  author={Honnibal, Matthew and Montani, Ines and Van Landeghem, Sofie and Boyd, Adriane},
  year={2020},
  doi={10.5281/zenodo.1212303}
}

@article{malkov2020hnsw,
  title={Efficient and Robust Approximate Nearest Neighbor Search Using Hierarchical Navigable Small World Graphs},
  author={Malkov, Yu A. and Yashunin, D. A.},
  journal={IEEE Transactions on Pattern Analysis and Machine Intelligence},
  volume={42},
  number={4},
  pages={824--836},
  year={2020},
  doi={10.1109/TPAMI.2018.2889473}
}

@article{douze2024faiss,
  title={The {Faiss} library},
  author={Douze, Matthijs and Guzhva, Alexandr and Deng, Chengqi and Johnson, Jeff and Szilvasy, Gergely and Mazar{\'e}, Pierre-Emmanuel and Lomeli, Maria and Hosseini, Lucas and J{\'e}gou, Herv{\'e}},
  journal={arXiv preprint arXiv:2401.08281},
  year={2024}
}
\bibliographystyle{iclr2027_conference}

\appendix
\newpage

\section{Implementation Details and Design Discussion}
\label{app:implementation}

The complete retrieval procedure of NexusRAG is given in Algorithm~\ref{alg:nexusrag}. The propagation state maintains a per-entity activation score $a_i^{(t)}$ for the current iteration. During each iteration, $\Tilde{a}_i^{(t)}$ records the semantic activation before structural expansion, while $a_i^{(t)}$ denotes the resulting activation after both paths. In contrast, the separate cumulative activation weight $A_i$ sums all admitted activation scores received by entity $e_i$ over the retrieval process. Thus, $a_i^{(t)}$ determines whether an entity can continue propagating, whereas $A_i$ determines how strongly the entity contributes to final passage retrieval. 

\begin{algorithm}[htbp]
\caption{NexusRAG Retrieval}
\label{alg:nexusrag}
\begin{algorithmic}[1]
\Require Corpus Tri-Graph $\gG$, query $q$, thresholds $\thresh,\delta$, max hops $maxT$, fusion coefficient $\alpha$, neighbor cap $\kappa$, top-$\eta$ sentences
\State Construct co-occurrence and semantic similarity score
\State Normalize nonzero co-occurrence and semantic similarity scores
\State Compute $w_{ij}$ using the normalized score using Eq.~\ref{eq:neighbor_fusion}
\State Construct $\matW$ by retaining neighbors with $w_{ij} \geq \thresh$ that rank within the top-$\kappa$ candidates for each entity
\State Apply NER and match each query entity to its most similar entity in $\gE$ to obtain seed set $E_0$
\State Initialize $a_i^{(0)}$ with the matching similarity for $e_i\in E_0$, and set all other entries to $0$
\State Initialize $A_i\gets a_i^{(0)}$ for each $e_i\in\gE$
\State Initialize $\mathrm{hop}(e_i)\gets1$ for each $e_i\in E_0$
\Statex \textbf{Dual-Path Entity Propagation}
\For{$t=1,\ldots,maxT$}
    \State Initialize $\Tilde{a}_j^{(t)}\gets0$ and $a_j^{(t)}\gets0$ for all $e_j\in\gE$
    \For{each active $e_i\in\gE$ with $a_i^{(t-1)}\geq\delta$} \Comment{Semantic propagation}
        \State Select top-$\eta$ query-similar sentences containing $e_i$
        \State Compute admitted semantic activations using Eq.~\ref{eq:hard_prop}
    \EndFor
    
    \For{each active $e_i\in\gE$ and $e_j\in\Neighbor(e_i)$}\Comment{Structural propagation}
        \State Compute the final activation $a_j^{(t)}$ using Eq.~\ref{eq:weak_prop}
    \EndFor
    \State Accumulate admitted activations into $A_j$ for each activated $e_j$
    \State Update $\mathrm{hop}(e_j)\gets t$ for each activated $e_j$
    \If{$a_j^{(t)}=0$ for all $e_j\in\gE$}
        \State \textbf{break}
    \EndIf
\EndFor
\State Initialize $P_{\mathrm{init}}(p_j)$ for each passage $p_j$ using similarity and accumulated activation score $A_i$ of entities \Comment{Eq.~\ref{eq:hybrid_init}}
\State Run PPR on the entity--passage subgraph using $P_{\mathrm{init}}(p_j)$ and the accumulated activation score
\State \Return Top-$K$ passages
\end{algorithmic}
\end{algorithm}

\subsection{Complexity}

Constructing sparse co-occurrence scores costs $O\!\big(\sum_{s \in \gS} |E_s|^2\big)$, linear in $|\gS|$ since each sentence $s$ holds $|E_s| \approx 4$ entities, where $E_s$ is the set of entities in $s$; the similarity term is built by ANN search instead of an all-pairs computation: building the Faiss (HNSW) index~\citep{douze2024faiss,malkov2020hnsw} costs $O(|\gE| \log |\gE| \cdot d_{\mathrm{emb}})$ and querying it for the top-$k$ neighbors of each entity costs $O(|\gE| \log |\gE| \cdot d_{\mathrm{emb}})$ in total (each ANN query is $O(\log |\gE| \cdot d_{\mathrm{emb}})$), materializing only $O(|\gE| \cdot k)$ similarity pairs, where $d_{\mathrm{emb}}$ is the embedding dimension. The resulting neighbor matrix $\matW$ is stored in CSR format with storage $O(|\gE| \cdot \kappa)$ after the top-$\kappa$ truncation. Per-query retrieval remains linear in the number of entities, matching LinearRAG's inference-time scaling, while the offline construction is reduced from quadratic to near-linear in $|\gE|$.

\subsection{Why Neighbors and Why Two Paths?}

The neighbor prior addresses the two failure modes and also affects answer localization.
\begin{itemize}[leftmargin=*]
 \item \textbf{Query-gated cutoff.} When the query is lexically distant from bridging sentences, the semantic propagation alone would assign minimal scores (capped at $\delta$), so genuine but weakly query-similar bridges are passed over. The structural propagation provides a fallback path through structural neighbors, ensuring that high-co-occurrence entities are not discarded due to surface-form mismatch.
 \item \textbf{Spurious activation.} Without a structural prior, a query-similar sentence can activate multiple co-occurring entities, including entities that are not on the relevant reasoning path. The neighbor clamp in Eq.~\ref{eq:hard_prop} bounds the transition by $w_{ij}$; non-neighbor transitions that pass the $\delta$ threshold receive the lower bound $\delta$, after which they cannot participate in the next propagation.
 \item \textbf{Answer localization.} For center-entity queries where the question revolves around a single entity, the answer is frequently found among that entity's contextual partners. The neighbor clamp in Eq.~\ref{eq:hard_prop} assigns non-neighbor transitions that pass the admission gate the minimum score $\delta$, so multi-source corroboration can still surface a genuinely co-occurring partner.
\end{itemize}

\textbf{Why operate on semantic activations?} Propagating from entities with nonzero $\Tilde{a}^{(t)}$ focuses structural expansion on entities newly activated by semantic propagation, allowing structural propagation to recover related entities that the semantic propagation may miss. This preserves query relevance from the semantic propagation while using the neighbor structure as a complementary expansion signal.

\textbf{Why two paths?} The semantic and structural propagations address two aspects of retrieval: the semantic propagation retains query-dependent sentence mediation while the neighbor weights limit transitions through structurally weak pairs (Section~\ref{sec:dual_path}); the structural propagation adds transitions along pre-computed neighbor edges without using sentence--query similarity. The structural propagation is seeded by semantic activation (Eq.~\ref{eq:weak_prop}), so it extends the frontier reached by sentence mediation rather than introducing an independent source of query entities. Together, the two paths suppress query-gated cutoff and spurious activation.

\subsection{Formal Properties of the Clamp and the structural propagation}
\label{app:properties}

\textbf{Property 1 (Terminal non-neighbor activation).}
For any admitted transition under Eq.~\ref{eq:hard_prop}, a neighbor receives a score bounded by its structural weight $w_{ij}$,
whereas a non-neighbor receives exactly the floor score $\delta$. Consequently, an admitted non-neighbor can contribute to the current evidence set and to the cumulative PPR weight, but it cannot obtain a propagation score above the floor and therefore cannot serve as an unconstrained source of further activation, except when \(\sigma_m=1\).

Under Eq.~\ref{eq:hard_prop}, a neighbor receives
\[
\max\!\left(\min(a_{i}^{(t-1)}\sigma_{m},w_{ij}),\delta\right),
\]
whereas a non-neighbor receives exactly $\delta$. Thus, non-neighbor transitions are restricted to the floor score and cannot obtain a higher propagation score through this branch, although their $\delta$ contributions remain available to the cumulative PPR weight $A_j$ and hence to passage initialization.

\textbf{Property 2 (Query-independent reachability).} For every entity with at least one co-occurring partner, normalization retains a neighbor with $w_{ij} \geq \alpha$, and that neighbor is activated by Eq.~\ref{eq:weak_prop} whenever the frontier entity's activation satisfies $\Tilde{a}_i^{(t)} \geq \delta/\alpha$, regardless of any query--sentence similarity.

The row maximum of the normalized co-occurrence term is $1$, so the fused weight of the maximum-weight co-occurrence partner is at least $\alpha$; this gives the $0\%$ zero-neighbor fraction whenever $\thresh \leq \alpha$ (Appendix~\ref{app:neighbor_stats}). The firing condition $w_{ij} \Tilde{a}_i^{(t)} \geq \delta$ of Eq.~\ref{eq:weak_prop} does not use query--sentence similarity, so the maximum-weight structural neighbor can remain reachable even when its bridging sentence has low query similarity.
 
\textbf{Remark (duality).} Query-gated cutoff and spurious activation are the false-negative and false-positive modes of the same admission decision. The two propagation paths address these failure modes in different ways: semantic propagation constrains sentence-mediated transitions using the corpus-level neighbor prior to reduce spurious activation, while structural propagation expands through structurally supported neighbors without relying on query-sentence similarity for the transition, helping recover potentially missed evidence.

\section{Datasets}
\label{app:datasets}

Our experimental evaluation is conducted on four datasets: three established multi-hop QA benchmarks, HotpotQA~\citep{yang2018hotpotqa}, 2WikiMultiHopQA (2Wiki)~\citep{2wikimqa}, and MuSiQue~\citep{trivedi2022musique}, plus one domain-specific Medical set from GraphRAG-Bench~\citep{xiang2025use}.

\textbf{(i) HotpotQA~\citep{yang2018hotpotqa}:} A benchmark comprising 97$k$ question-answer instances designed to evaluate multi-hop reasoning capabilities. Each question requires models to synthesize information from multiple documents, with up to 2 gold-standard supporting passages provided alongside numerous irrelevant documents.

\textbf{(ii) 2WikiMultiHopQA (2Wiki)~\citep{2wikimqa}:} A multi-hop reasoning benchmark containing 192$k$ questions that necessitate information integration across multiple Wikipedia articles. Each instance requires evidence synthesis from either 2 or 4 specific articles, testing models' ability to perform structured cross-document reasoning and maintain coherent information flow.

\textbf{(iii) MuSiQue~\citep{trivedi2022musique}:} A multi-hop QA benchmark featuring 25$k$ question-answer pairs that demand 2-4 sequential reasoning steps. Each question requires coherent multi-step logical inference across multiple documents.

\textbf{(iv) Medical:} A specialized subset derived from GraphRAG-Bench~\citep{xiang2025use}, constructed from structured clinical data sourced from the National Comprehensive Cancer Network (NCCN) guidelines. These guidelines provide standardized treatment protocols, drug interaction hierarchies, and diagnostic criteria. The dataset encompasses four tasks of increasing complexity: fact retrieval, complex reasoning, contextual summarization, and creative generation. GraphRAG-Bench spans 4,076 questions across these difficulty levels in total; the Medical subset used in our evaluation comprises 2,062 questions.

\section{Baseline Details}
\label{app:baselines}

We compare NexusRAG against the same set of retrieval and GraphRAG baselines as LinearRAG~\citep{zhuang2025linearrag}, plus LinearRAG itself as the direct baseline.

\textbf{(i) Vanilla RAG} retrieves the top-$K$ passages by dense similarity and feeds them directly to the generator, providing a retrieval-augmented lower bound without any graph structure.

\textbf{(ii) KGP}~\citep{wang2024knowledge} builds a knowledge graph over multiple passages, with edges encoding their semantic and lexical similarity; at retrieval, it introduces an LLM-driven graph-traversal agent to navigate the graph and progressively collect supporting passages.

\textbf{(iii) G-Retriever}~\citep{he2024g} combines graph neural networks with LLMs by formulating subgraph retrieval as a Prize-Collecting Steiner Tree optimization problem, for conversational question answering on textual graphs while mitigating hallucination and enhancing scalability.

\textbf{(iv) RAPTOR}~\citep{sarthi2024raptor} builds a hierarchical tree by applying clustering algorithms and abstractive summarization, facilitating representation at multiple semantic granularities.

\textbf{(v) E2GraphRAG}~\citep{zhao20252graphrag} uses spaCy to extract entities and LLMs to summarize passage groups into a hierarchical tree with encoded nodes at indexing; at retrieval, it hits relevant entities in their multi-hop neighborhood and collects associated passages for ranking, otherwise performing dense retrieval over the whole tree.

\textbf{(vi) LightRAG}~\citep{guo2024lightrag} employs a two-tier framework that incorporates graph-based representations within textual indexing, merging fine-grained entity--relation mappings with coarse-grained thematic structures.

\textbf{(vii) HippoRAG}~\citep{HippoRAG} is a training-free graph-enhanced retriever that uses the Personalized PageRank algorithm with query concepts as seeds for single-step or multi-hop retrieval across disparate documents.

\textbf{(viii) GFM-RAG}~\citep{luo2025gfm} implements a GraphRAG paradigm by constructing graphs from documents and using a graph-enhanced retriever to retrieve relevant documents.

\textbf{(ix) HippoRAG2}~\citep{gutiérrez2025hipporag2} extends HippoRAG with enhanced paragraph integration and contextualization, optimizing seed-node selection and PageRank reset probabilities while maintaining factual-memory capabilities.

\textbf{(x) LinearRAG}~\citep{zhuang2025linearrag} is the direct baseline used for comparison; it performs linear-time retrieval over an entity--sentence--passage tri-graph via Personalized PageRank, without the entity-neighbor modeling introduced in this work.

\section{Machine Configuration}
\label{app:machine}

Retrieval and indexing experiments were conducted on the hardware in Table~\ref{tab:machine_config}. Generation and evaluation were performed using three setups: GPT-4o-mini and DeepSeek-V4-Flash via their respective APIs, with Qwen3.6-27B-FP8 running on the same hardware below.

\begin{table}[h]
\centering
\caption{{Detailed machine configuration used in our experiments.}}
\label{tab:machine_config}
\begin{tabular}{lc}
\toprule
Component & Specification \\
\midrule
GPU & NVIDIA RTX A6000 \\
CPU & Intel(R) Xeon(R) Platinum 8260 CPU @ 2.30GHz \\
CUDA & 12.8 (Driver 570.172.08) \\
\bottomrule
\end{tabular}
\vspace{-2mm}
\end{table}

\section{Hyperparameter Configurations}
\label{app:hyperparams}

Table~\ref{tab:hyperparams} reports the full hyperparameter configuration. The propagation and PPR parameters $maxT, \delta, \eta, d, \omega_p$ are inherited from LinearRAG~\citep{zhuang2025linearrag} and held fixed across all datasets. The neighbor-mechanism parameters are set as: fusion coefficient $\alpha = 0.5$, neighbor threshold $\thresh = 0.5$, and maximum neighbors per entity $\kappa = 5$. The neighbor cap $\kappa = 5$ bounds the per-entity degree to limit memory and propagation compute; its role as a recall-versus-cost tradeoff is characterized in Section~\ref{sec:param_sensitivity}.

\begin{table}[h]
\centering
\caption{{Hyperparameter configuration.} $maxT$: max propagation hops; $\delta$: pruning threshold; $\eta$: sentences per entity per iteration; $d$: PPR damping; $\omega_p$: passage node weight. The first group is fixed to LinearRAG's reported values; $\alpha, \thresh, \kappa$ are the neighbor-mechanism parameters (the structural propagation uses the raw neighbor weight $w_{ij}$ with no extra decay coefficient).}
\label{tab:hyperparams}
\small
\begin{tabular}{lcccccccc}
\toprule
\textbf{Dataset} & $maxT$ & $\delta$ & $\eta$ & $d$ & $\omega_p$ & $\alpha$ & $\thresh$ & $\kappa$ \\
\midrule
HotpotQA & 3 & 0.4 & 1 & 0.5 & 0.05 & 0.5 & 0.5 & 5 \\
2Wiki & 3 & 0.4 & 1 & 0.5 & 0.05 & 0.5 & 0.5 & 5 \\
MuSiQue & 5 & 0.1 & 4 & 0.5 & 0.05 & 0.5 & 0.5 & 5 \\
Medical & 3 & 0.5 & 3 & 0.5 & 0.05 & 0.5 & 0.5 & 5 \\
\bottomrule
\end{tabular}
\vspace{-2mm}
\end{table}

\section{Neighbor Matrix Statistics}
\label{app:neighbor_stats}

\textbf{Remark (inclusive neighbor construction).} Threshold-based admission has a bounded form: high-weight neighbors retain their transition scores up to their neighbor weights, low-weight edges are admitted with attenuation, and non-neighbors fall to the lower bound $\delta$ (Section~\ref{sec:dual_path}). In particular, when $\thresh \leq \alpha$ every entity retains at least its maximum-weight co-occurrence partner; normalization guarantees $w_{ij} \geq \alpha \geq \thresh$ for the row maximum, so before rank truncation, every entity with at least one co-occurring partner has at least one candidate with fused weight at least $\alpha$. At the adopted $\kappa=5$ and $\thresh=0.5$, the resulting effective graph empirically has 0

\begin{table}[t]
\vspace{-2mm}
\centering
\caption{{Neighbor graph statistics on the Qwen3.6-27B-FP8 backbone, after the top-$\kappa{=}5$ truncation (the effective graph used by propagation).} Avg.\ \#neighbors = average neighbors per entity in the effective graph; Zero-neighbor (\%) = fraction of entities with no effective neighbors. The top half reports per-dataset statistics at the adopted threshold $\thresh=0.5$; the bottom half reports per-threshold statistics on HotpotQA.}
\label{tab:neighbor_stats}
\setlength{\tabcolsep}{2mm}
\begin{tabular}{lccc}
\toprule
\textbf{Dataset} & \textbf{$\thresh$} & \textbf{Avg.\ \#neighbors} & \textbf{Zero-neighbor (\%)} \\
\midrule
\multirow{4}{*}{Per dataset at $\thresh=0.5$ (adopted)}
& HotpotQA & 1.99 & 0.00 \\
& 2Wiki & 1.99 & 0.00 \\
& MuSiQue & 2.01 & 0.00 \\
& Medical & 2.29 & 0.00 \\
\midrule
\multirow{6}{*}{Per threshold on HotpotQA}
& 0.4 & 4.38 & 0.00 \\
& 0.5 & 1.99 & 0.00 \\
& 0.6 & 0.03 & 97.54 \\
& 0.7 & 0.01 & 99.02 \\
& 0.8 & 0.01 & 99.23 \\
& 0.9 & 0.01 & 99.46 \\
\bottomrule
\end{tabular}
\vspace{-2mm}
\end{table}

Table~\ref{tab:neighbor_stats} reports neighbor graph statistics for the Qwen3.6-27B-FP8 backbone after the top-$\kappa{=}5$ truncation, i.e., the effective graph that the propagation rule actually uses; we report per-dataset statistics at the adopted $\thresh=0.5$ and per-threshold statistics on HotpotQA for $\thresh \in \{0.4, 0.5, 0.6, 0.7, 0.8, 0.9\}$. Higher $\thresh$ yields sparser graphs and a gentle, near-monotone accuracy decline: at $\thresh \ge 0.6$ the zero-neighbor fraction on HotpotQA jumps to $97$--$99\%$ and LLM-Acc.\ falls from $88.5$ to $87.9$, whereas the denser $\thresh=0.4$ graph ($4.38$ neighbors per entity) is the most accurate ($89.0$). The $\sim\!66\times$ density collapse between $\thresh=0.5$ and $\thresh=0.6$ costs only $0.2$ points, so those removed edges are low-weight and rarely firing; the further $0.6$ points lost up to $\thresh=0.9$ come from the higher-weight edges that remain. This is expected: the neighbor graph is a prior on \emph{plausible} partners, not all of which matter for a given query, so accuracy need not track edge density; distant low-weight partners can be deleted almost for free, while the few higher-weight partners that remain are more likely to fire, so their removal costs visibly more. This also reconciles the sweep with the ablation in Section~\ref{sec:ablation}: the contribution concentrates in the few higher-weight edges, which is why removing almost all low-weight edges costs only $0.2$ points while removing the mechanism entirely costs several. We adopt $\thresh = \alpha = 0.5$ : every entity retains its maximum-weight co-occurrence partner by construction, low-weight edges are admitted and self-attenuate by their weight, and the average neighbor count of $1.99$--$2.29$ stays well below the cap, so $\kappa{=}5$ rarely truncates.

\section{Detailed Parameter Sensitivity}
\label{app:param_detail}

\subsection{Impact of Fusion Coefficient $\alpha$ and Neighbor Threshold $\thresh$}

\begin{figure}[htbp]
 \centering
 \begin{subfigure}[t]{0.48\linewidth}
 \includegraphics[width=\linewidth]{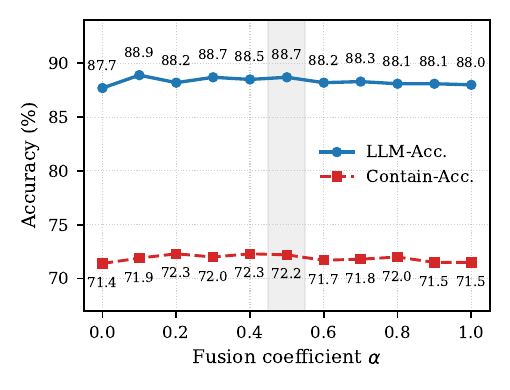}
 \caption{Sensitivity to $\alpha$.}
 \end{subfigure}
 \hfill
 \begin{subfigure}[t]{0.48\linewidth}
 \includegraphics[width=\linewidth]{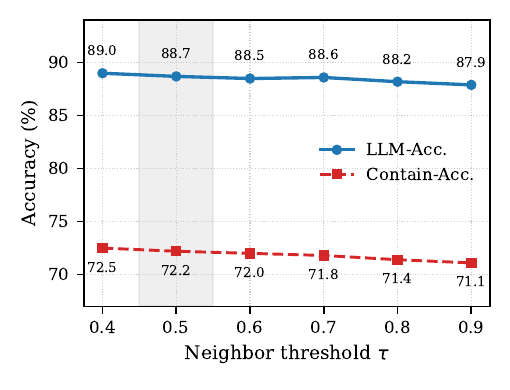}
 \caption{Sensitivity to $\thresh$.}
 \end{subfigure}
 \caption{{Sensitivity to the fusion coefficient $\alpha$ and the neighbor threshold $\thresh$ on the Qwen3.6-27B-FP8 backbone (HotpotQA; one parameter varied at a time, the other fixed at its adopted value).}}
 \label{fig:alpha_tau_sensitivity}
 \vspace{-4mm}
\end{figure}

We first examine the fusion coefficient $\alpha$ and the neighbor threshold $\thresh$, sweeping one parameter at a time on HotpotQA with the Qwen3.6-27B-FP8 backbone and the other fixed at its adopted value (Figure~\ref{fig:alpha_tau_sensitivity}); $\alpha = \thresh = 0.5$ is fixed for all datasets.

\obs Across the swept $\alpha$ range $[0.0,1.0]$ at $\thresh=0.5$, LLM-Acc.\ varies by at most $1.2$ points; within the fusion range $\alpha \in [0.1,0.9]$ it stays at $88.1$--$88.9$ (Contain-Acc.\ $71.5$--$72.3$), while the pure boundaries $\alpha=0$ ($87.7$/$71.4$) and $\alpha=1$ ($88.0$/$71.5$, touching the Contain-Acc.\ lower edge) sit at or just below that band, so each single signal retains most of the benefit yet never exceeds the fused interior. Across the swept $\thresh$ range $[0.4,0.9]$ at $\alpha=0.5$, LLM-Acc.\ declines gently from $89.0$ at $\thresh=0.4$ to $87.9$ at $\thresh=0.9$ ($88.7$, $88.5$, $88.6$, $88.2$ at $\thresh=0.5$, $0.6$, $0.7$, $0.8$), a total range of $1.1$ points, while Contain-Acc.\ is highest at the dense $\thresh=0.4$ setting ($72.5$) and otherwise stays within $71.1$--$72.2$.

An edge fires only when $\Tilde{a}_i^{(t)} \geq \delta / w_{ij}$, so a low-weight edge contributes little while present: raising $\thresh$ from $0.5$ to $0.6$ removes mostly such edges, collapsing the graph from $1.99$ to $0.03$ neighbors per entity ($\sim\!66\times$) for $0.2$ LLM-Acc.\ points. From $\thresh=0.6$ to $0.9$ the graph holds $0.03$ to $0.01$ neighbors per entity, yet accuracy falls a further $0.6$ points, so the edges remaining at the highest thresholds account for most of the contribution. Semantic propagation continues under the $\delta$ lower bound throughout. We adopt $\thresh = \alpha = 0.5$, which retains the maximum-weight co-occurrence partner of every entity by construction. The ablation in Section~\ref{sec:ablation} establishes the same point independently: removing the neighbor mechanism degrades accuracy, showing that both signals contribute.

\subsection{Impact of Neighbor Cap $\kappa$}
The neighbor cap $\kappa$ bounds the per-entity degree of the effective graph and thereby limits the neighbor-matrix memory footprint and the per-iteration propagation work ($O(|\gE|\cdot\kappa)$, with the multi-hop reachable set scaling with $\kappa$ as a branching factor).
At the adopted $\thresh=0.5$ the effective graph averages only $1.99$--$2.29$ neighbors per entity (Table~\ref{tab:neighbor_stats}), well below the cap, so $\kappa=5$ rarely truncates in practice: it serves as a hard upper bound on memory and propagation cost rather than as an active recall parameter.

\begin{wrapfigure}{r}{0.46\linewidth}
 \centering
 \vspace{-4mm}
 \includegraphics[width=\linewidth]{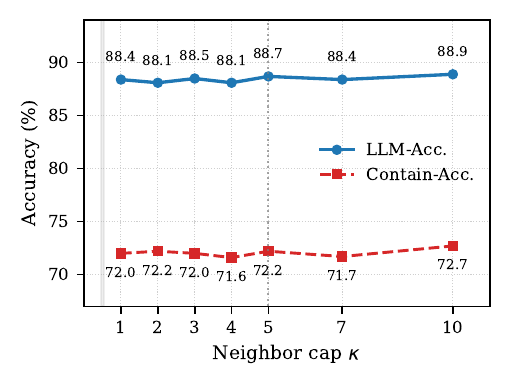}
 \caption{Sensitivity to the neighbor cap $\kappa$.}
 \vspace{-4mm}
 \label{fig:kappa_sensitivity}
\end{wrapfigure}

We verify this directly by sweeping $\kappa \in \{1,2,3,4,5,7,10\}$ on HotpotQA (Qwen3.6-27B-FP8), holding $\alpha=\thresh=0.5$ fixed (Figure~\ref{fig:kappa_sensitivity}).

\obs Over $\kappa = 1, 2, 3, 4, 5, 7, 10$, LLM-Acc.\ takes the values $88.4$, $88.1$, $88.5$, $88.1$, $88.7$, $88.4$, $88.9$ and Contain-Acc.\ $72.0$, $72.2$, $72.0$, $71.6$, $72.2$, $71.7$, $72.7$; the adopted $\kappa=5$ setting ($88.7$/$72.2$, from Table~\ref{tab:main_results}) falls within this $0.8$-point range. The variation is non-monotonic rather than trended: the largest cap $\kappa=10$ gives the largest value in this sweep yet exceeds the adopted setting by only $0.2$ LLM-Acc.\ points, while $\kappa=2$ and $\kappa=4$ give the lowest, $0.6$ points below it. This is consistent with the effective graph averaging only $1.99$--$2.29$ neighbors per entity, so a cap as small as $\kappa=3$ already rarely truncates and $\kappa$ bounds compute rather than recall; $\kappa=5$ lies inside this flat region and is fixed across all datasets.

\section{Retrieval-level Recall Evaluation}
\label{app:retrieval_recall}

We report retrieval-level recall on the two benchmarks with annotated bridge evidence. 
Table~\ref{tab:recall_at_k} reports Answer-Recall@$K$ and Bridge-Recall@$K$ at two retrieval depths. The structural propagation increases Answer-Recall at both depths on both datasets, with gains up to $2.0$ points. Bridge-Recall also increases, with gains of $2.27$/$3.51$ points on HotpotQA and $0.69$/$1.53$ points on 2WikiMultiHopQA at $K{=}2/5$, respectively, providing a finer-grained view of evidence-chain retrieval.

\begin{wraptable}{r}{0.5\linewidth}
\centering
\centering
\caption{{Retrieval-level recall@$K$ (\%).} Bridge-Recall (Br) and Answer-Recall (Ans) at each depth $K$. Highest values are in \textbf{bold}.}
\vspace{-1mm}
\label{tab:recall_at_k}
\setlength{\tabcolsep}{1mm}
\resizebox{\linewidth}{!}{%
\begin{tabular}{lcccccccc}
\toprule
\multirow{3}{*}{\textbf{Method}}
 & \multicolumn{4}{c}{\textbf{HotpotQA}}
 & \multicolumn{4}{c}{\textbf{2Wiki}}\\
 \cmidrule(lr){2-5} \cmidrule(lr){6-9}
 & \multicolumn{2}{c}{$K{=}2$}
 & \multicolumn{2}{c}{$K{=}5$}
 & \multicolumn{2}{c}{$K{=}2$}
 & \multicolumn{2}{c}{$K{=}5$}\\
 \cmidrule(lr){2-3} \cmidrule(lr){4-5} \cmidrule(lr){6-7} \cmidrule(lr){8-9}
 & Br & Ans & Br & Ans & Br & Ans & Br & Ans \\
\midrule
LinearRAG & 87.71 & 78.1 & 89.98 & 82.8 & 80.36 & 72.0 & 81.6 & 76.3 \\
NexusRAG & \textbf{89.98} & \textbf{78.7} & \textbf{93.49} & \textbf{84.8} & \textbf{81.05} & \textbf{72.7} & \textbf{83.13} & \textbf{76.7} \\
\bottomrule
\end{tabular}}
\vspace{-5mm}
\end{wraptable}

Finally, the retrieval difference is also reflected in generation: the judge-independent Contain-Acc.\ rises in the same direction (HotpotQA +4.6pp, 2Wiki +2.7pp, Table~\ref{tab:main_results}), and on the Qwen backbones NexusRAG's Contain-Acc.\ exceeds its own Answer-Recall on 2WikiMultiHopQA (81.4 vs.\ 76.7; Table~\ref{tab:main_results} and Table~\ref{tab:recall_at_k}), showing that the retrieval difference remains visible in the generated answers.

\section{Case Studies: Two Failure Modes of Query-Gated Propagation}
\label{app:case_study}

We concretize the two failure modes of Section~\ref{sec:preliminary} with two real multi-hop instances on which NexusRAG answers correctly and LinearRAG does not. In each case the gold reasoning chain is shown as \emph{Support Context} and the two systems' retrieved context is compared side by side (Tables~\ref{tab:case_study} and~\ref{tab:case_spurious}). Case~\ref{app:case_cutoff} illustrates \emph{query-gated cutoff} in a representative case: LinearRAG never anchors the bridge entity, so the answer-bearing passage is missing from its context entirely. Case~\ref{app:case_spurious} illustrates \emph{spurious activation}: LinearRAG does complete the first hop, but the bridging sentence names two co-occurring entities and the walk moves to the other co-occurring entity, so the second hop terminates on the wrong branch and the answer passage is again never retrieved. In both cases, NexusRAG recovers a passage that LinearRAG does not return.

\begin{table*}[h]
\caption{{Case study (HotpotQA).} LinearRAG retrieves topically related poet / civil-rights passages, one of which mentions Maya Angelou only in passing, but misses the bridge article \emph{Ford Hall Forum} (which contains the answer sentence) and answers incorrectly, whereas NexusRAG's structural propagation retains the \emph{Ford Hall Forum} passage and answers correctly.}
\label{tab:case_study}
\centering
\resizebox{0.98\textwidth}{!}{%
\begin{tabular}{p{3cm}|p{13cm}}
\toprule
\textbf{Question} & ``What is unique about the forum an American poet, memoirist, and civil rights activist spoke at?'' \\
\midrule
\textbf{Ground Truth} & the oldest free public lecture series in the United States \\
\midrule
\textbf{Support Context} & [``American poet, memoirist \& civil-rights activist''] $\to$ ``Maya Angelou'' $\to$ ``spoke at Ford Hall Forum'' \\
& [``Ford Hall Forum''] $\to$ ``the oldest free public lecture series in the United States'' \\
\midrule
\textbf{LinearRAG} & \textbf{\textit{Retrieved context:}} \\
& 1) \ding{53} ``Cleanth Brooks (poet \& critic)'': \ldots with \emph{The Southern Review} in 1935; won the Pulitzer Prize for fiction and for poetry \ldots \\
& 2) \ding{53} ``Grosvenor, Duke of Westminster'': \ldots arts \& charity patron; mentions Maya Angelou only in passing \ldots \\
& 3) \ding{53} ``Derick Wade Burleson (poet)'': \ldots American academic and writer; \emph{Ejo: Poems, Rwanda 1991--94} \ldots \\
& 4) \ding{53} ``Beamus Pierce / Eagle Feather (Smithsonian trustee)'': \ldots board of trustees for the Smithsonian's National Museum of the American Indian \ldots \\
& 5) \ding{53} ``American Pit Bull Terrier'': \ldots competes in dog sports and conformation shows \ldots \\
& \emph{(The bridge article \emph{Ford Hall Forum} is absent from the retrieved context.)} \\
& \textbf{\textit{Prediction:}} \\
& \ding{53} It is the United Nations. \\
\midrule
\textbf{NexusRAG (ours)} & \textbf{\textit{Retrieved context:}} \\
& 1) \ding{51} ``Ford Hall Forum (in a civil-rights biography)'': \ldots \emph{the Ford Hall Forum is the oldest free public lecture series in the United States} \ldots past speakers include Maya Angelou, Noam Chomsky, Martin Luther King Jr. \ldots \\
& 2) \ding{53} ``Cleanth Brooks (poet \& critic)'': \ldots with \emph{The Southern Review} in 1935; won the Pulitzer Prize for fiction and for poetry \ldots \\
& 3) \ding{53} ``Derick Wade Burleson (poet)'': \ldots American academic and writer; \emph{Ejo: Poems, Rwanda 1991--94} \ldots \\
& 4) \ding{53} ``Langston Hughes gospel show'': \ldots traditional Christmas carols are sung in gospel style \ldots originally written by Langston Hughes \ldots \\
& 5) \ding{53} ``Beamus Pierce / Eagle Feather (Smithsonian trustee)'': \ldots board of trustees for the Smithsonian's National Museum of the American Indian \ldots \\
& \textbf{\textit{Prediction:}} \\
& \ding{51} It is the oldest free public lecture series in the United States. \\
\cr \bottomrule
\end{tabular}}
\vspace{-4mm}
\end{table*}

\subsection{Case 1: Query-Gated Cutoff Drops the Bridge}
\label{app:case_cutoff}

The question asks what is distinctive about a forum addressed by an American poet, memoirist, and civil-rights activist. Its gold reasoning chain is two-hop: the activist resolves to \emph{Maya Angelou}, who spoke at the \emph{Ford Hall Forum}, and that forum is defined as the oldest free public lecture series in the United States. The relevant bridge is therefore the entity \emph{Ford Hall Forum}, whose defining sentence (``the oldest free public lecture series'') is lexically distant from the query's ``American poet, memoirist, and civil rights activist''.

LinearRAG answers incorrectly. Its retriever returns several topically adjacent poet / civil-rights passages, including one that mentions Maya Angelou only in passing (the Grosvenor biography), but never surfaces the \emph{Ford Hall Forum} article that contains the answer sentence. Because the answer-bearing passage is absent from the context, the generator cannot recover it and instead outputs the unrelated ``United Nations''. This is an instance of \emph{query-gated cutoff} (Section~\ref{sec:preliminary}): the query-gated propagation fails to carry activation across the weakly query-similar bridge to \emph{Ford Hall Forum}.

NexusRAG answers correctly. In addition to the query-gated sentence-mediated propagation, its neighbor mechanism performs a \emph{structural} expansion (Eq.~\ref{eq:weak_prop}) along pre-computed entity--entity edges that are sentence-independent. This expansion reaches the weakly query-similar bridge \emph{Ford Hall Forum} before the query-gated propagation removes it from the final context. The case is consistent with the role of structural propagation in preserving a weakly query-similar bridge during subsequent passage ranking. The generator then reads off ``the oldest free public lecture series in the United States.'' This illustrates the role of the structural propagation in retrieving bridges whose sentences have low query similarity.

\subsection{Case 2: Spurious Activation Diverts the Second Hop}
\label{app:case_spurious}

The second instance is a 2WikiMultihopQA question on which LinearRAG \emph{does} complete the first hop and retrieves the bridging passage, yet still fails, even though it names the correct bridge entity in its own output, because the second hop moves to an entity that merely co-occurs with the bridge in the same sentence. The two systems share four of their five retrieved passages; Table~\ref{tab:case_spurious} compares the two contexts.

\begin{table*}[h]
\caption{{Case study (2WikiMultiHopQA).} LinearRAG retrieves the hop-1 bridging passage (349) and correctly identifies the mother, \emph{Infanta Blanca of Spain}, but its second-hop expansion follows the \emph{father} \emph{Leopold Salvator}, the entity that co-occurs with the mother in that very sentence, and returns his biography (630) instead; the mother's own article (147), which states her date of birth, is never retrieved, and LinearRAG answers ``not mentioned in the text''. NexusRAG returns the same hop-1 passage \emph{and} retains passage 147, and answers correctly. Passage IDs are the retriever's own indices; \ding{51} marks passages that carry a gold fact.}
\label{tab:case_spurious}
\centering
\resizebox{0.98\textwidth}{!}{%
\begin{tabular}{p{3cm}|p{13cm}}
\toprule
\textbf{Question} & ``What is the date of birth of Archduke Karl Pius of Austria, Prince of Tuscany's mother?'' \\
\midrule
\textbf{Ground Truth} & 7 September 1868 \\
\midrule
\textbf{Support Context} & [``Archduke Karl Pius of Austria, Prince of Tuscany''] $\xrightarrow{\text{\;mother\;}}$ ``Infanta Blanca of Spain'' \\
& [``Infanta Blanca of Spain''] $\xrightarrow{\text{\;date of birth\;}}$ ``7 September 1868'' \\
\midrule
\textbf{LinearRAG} & \textbf{\textit{Retrieved context:}} \\
& 1) \ding{51} \emph{hop-1 bridge} (passage 349): ``\ldots archduke \emph{karl pius} of austria, prince royal of hungary and bohemia, prince of tuscany (4 december 1909 -- 24 december 1953) \ldots he was the tenth and youngest child of archduke \emph{leopold salvator}, prince of tuscany \emph{and infanta blanca of spain}.'' \\
& 2) \ding{53} Habsburg / film biography chunk (passage 348). \\
& 3) \ding{53} \emph{wrong branch} (passage 630, retrieved only by LinearRAG): ``archduke \emph{leopold salvator}, prince of tuscany \ldots (\emph{15 october 1863} -- 4 september 1931), was the son of archduke karl salvator of austria \ldots''; the \emph{father}'s article; it contains a date of birth, but not the mother's. \\
& 4) \ding{53} Bolivian-mountains / film-director chunk (passage 350). \\
& 5) \ding{53} Habsburg princesses chunk (passage 148): names ``\emph{infanta blanca of spain}'' again, but only as a parent, with no date. \\
& \emph{(The article of \emph{Infanta Blanca of Spain}, which carries ``7 september 1868'', is absent from the retrieved context.)} \\
& \textbf{\textit{Prediction:}} \\
& \ding{53} ``\ldots the text lacks the info \ldots the father's birth date is \emph{15 October 1863}, but that is factually incorrect for the mother \ldots I will stick with `Not mentioned in the text'.'' \\
\midrule
\textbf{NexusRAG (ours)} & \textbf{\textit{Retrieved context:}} \\
& 1) \ding{51} \emph{hop-1 bridge} (passage 349): the same passage as above, establishing \emph{Karl Pius} $\to$ mother \emph{Infanta Blanca of Spain}. \\
& 2)--4) \ding{53} the same three distractors (passages 348, 350, 148). \\
& 5) \ding{51} \emph{hop-2 answer passage} (passage 147, returned only by NexusRAG): ``\emph{infanta blanca of spain} (\emph{7 september 1868} -- 25 october 1949) was the eldest child of carlos, duke of madrid \ldots in 1889 she married archduke leopold salvator of austria.'' \\
& \textbf{\textit{Prediction:}} \\
& \ding{51} 7 September 1868. \\
\cr \bottomrule
\end{tabular}}
\vspace{-4mm}
\end{table*}

The question is two-hop: \emph{Karl Pius} resolves to his mother \emph{Infanta Blanca of Spain}, and her article gives the date of birth. Both hops hinge on a single sentence in passage 349, which states that Karl Pius ``was the tenth and youngest child of archduke \emph{Leopold Salvator} \ldots \emph{and infanta Blanca of Spain}''; that is, the sentence names \emph{both} parents, only one of whom is the mother the query asks for.

LinearRAG answers incorrectly, and its own output shows where the chain breaks. It retrieves passage 349, and its generation explicitly identifies ``\emph{Archduke Leopold Salvator} (1863--1931), he is the \emph{father}'' and searches for ``Blanca'' with a date, so hop-1 succeeded. What it never retrieves is the mother's own article. Its only passage not shared with NexusRAG is passage 630, the biography of \emph{Leopold Salvator}: the second-hop expansion followed an entity that shares the bridging sentence rather than the queried one. This is \emph{Spurious activation} (Section~\ref{sec:preliminary}): a query-similar sentence activates \emph{every} entity it contains, and with no structural prior to tell which co-occurrence realizes the queried relation, the walk assigns higher activation to the wrong parent. Its downstream consequence is a retrieval gap: the answer passage is missing, so the generator can only report that the date is ``not mentioned in the text'' (it even notes the father's 15 October 1863 as a tempting but wrong substitute).

\begin{wraptable}{r}{0.48\linewidth}
\vspace{-3mm}
\centering
\caption{{Indexing and retrieval timing on the four benchmarks (seconds).} }
\label{tab:effbench_timing}
\scriptsize
\setlength{\tabcolsep}{1.2mm}
\begin{tabular}{lcccc}
\toprule
& \textbf{HotpotQA} & \textbf{2Wiki} & \textbf{MuSiQue} & \textbf{Medical} \\
\midrule
\multicolumn{5}{l}{\textbf{LinearRAG}} \\
\quad Index(s) & 1057.42 & 541.67 & 1109.22 & 179.86 \\
\quad Precompute(s) & 0.00 & 0.00 & 0.00 & 0.00 \\
\quad Retrieval(s) & 0.252 & 0.214 & 0.240 & 0.216 \\
\multicolumn{5}{l}{\textbf{NexusRAG (ours)}} \\
\quad Index(s) & 991.31 & 549.20 & 1072.60 & 181.85 \\
\quad Precompute(s) & 24.87 & 12.95 & 23.96 & 1.91 \\
\quad Retrieval(s) & 0.260 & 0.213 & 0.255 & 0.216 \\
\bottomrule
\end{tabular}
\vspace{-2mm}
\end{wraptable}

NexusRAG answers correctly, and the decisive element is the semantic propagation's neighbor clamp (Eq.~\ref{eq:hard_prop}, Section~\ref{sec:method}). A transition from an activated entity is admitted only up to the weight of the pre-computed neighbor edge, so an entity that merely shares the bridging sentence is capped rather than freely inherited. \emph{Leopold Salvator} (the father) is not in the neighbor set of \emph{Karl Pius} at the adopted setting (top-$\kappa{=}5$; Table~\ref{tab:neighbor_stats}), so the sentence-mediated path propagating through the shared sentence of passage~349 assigns him the lower bound $\delta$ instead of the full transition score; his biography (passage~630) therefore does not enter the top-5 through this transition. Because the mother's article (147), which carries the date of birth, already sits at the edge of the dense retriever's ranking, it is retained once the incidental father is no longer amplified, and the generator reads off ``7 September 1868''. The retrieval difference is consistent with the lower activation assigned to the incidental father under the structural propagation.

\section{Efficiency and Scalability}
\label{app:efficiency}

We also measure the runtime overhead introduced by the additional neighbor structure, since graph-based propagation can be slow at serving time. Table~\ref{tab:effbench_timing} reports the per-benchmark timing to address this: it separates the one-time preprocessing cost from the base graph index cost and from the average online retrieval cost (mean wall-clock time per sample over the test split), to separate the one-time preprocessing cost from the base indexing and online retrieval costs.

The measurements show that the additional cost is concentrated in preprocessing. The one-time \texttt{Precompute(s)} for NexusRAG ranges from about $2$\,s on Medical to roughly $25$\,s on HotpotQA and MuSiQue, while LinearRAG has no neighbor stage. The base \texttt{Index(s)} differs by only a few percent between the two methods, consistent with their shared graph index construction. \texttt{Retrieval(s)} differs by a few percent on three benchmarks and by about 6\% on MuSiQue (e.g., $0.260$\,s vs $0.252$\,s on HotpotQA, $0.216$\,s vs $0.216$\,s on Medical). The structural propagation reuses the cached neighbor matrix and adds a vectorized step. The reported results therefore add a one-time preprocessing stage without changing the basic online retrieval procedure.

We now extend this per-benchmark timing study to the large-scale ATLAS-Wiki corpus, using the 5M and 10M token subsets introduced by LinearRAG~\citep{zhuang2025linearrag}.

\begin{table}[ht]
\centering
\caption{{Indexing cost on ATLAS-Wiki (5M and 10M tokens).} \texttt{Total Index (s)} = base graph index (\texttt{Index(s)}) + neighbor precompute (\texttt{Precompute(s)}). LinearRAG has no neighbor precompute stage, so its total equals \texttt{Index(s)}; NexusRAG adds the one-time neighbor construction. HippoRAG figures are measured in this work (Qwen3.6-27B-FP8 via vLLM).}
\label{tab:large_scale_index}
\setlength{\tabcolsep}{1.8mm}
\resizebox{\columnwidth}{!}{%
\begin{tabular}{lcccccc}
\toprule
\multirow{2}{*}{\textbf{Dataset}} & \multirow{2}{*}{\textbf{Method}} & \multirow{2}{*}{\textbf{Index(s)}} & \multirow{2}{*}{\textbf{Precompute(s)}} & \multirow{2}{*}{\textbf{Total Index (s)}} & \multicolumn{2}{c}{\textbf{Token ($\times10^6$)}} \\
\cmidrule(lr){6-7}
& & & & & \textbf{Prompt} & \textbf{Completion} \\
\midrule
\multirow{3}{*}{5M}
& HippoRAG & 71019.97 & N/A & 71019.97 & 18.14 & 9.63 \\
& LinearRAG & 4188.49 & N/A & 4188.49 & 0 & 0 \\
& NexusRAG (ours) & 3949.79 & 106.93 & 4056.72 & 0 & 0 \\
\midrule
\multirow{3}{*}{10M}
& HippoRAG & 100245.49 & N/A & 100245.49 & 37.10 & 20.26 \\
& LinearRAG & 8110.30 & N/A & 8110.30 & 0 & 0 \\
& NexusRAG (ours) & 8077.13 & 228.87 & 8306.00 & 0 & 0 \\
\bottomrule
\end{tabular}%
}
\end{table}

Table~\ref{tab:large_scale_index} reports the one-time indexing cost of a complete rebuild from scratch (caches cleared before each run). LinearRAG has no neighbor precompute stage, so its total equals \texttt{Index(s)}, while NexusRAG adds the one-time neighbor construction that enables the structural propagation; HippoRAG likewise has no precompute stage, so its total equals its indexing time (measured here with Qwen3.6-27B-FP8 via vLLM). The prompt and completion tokens denote LLM input/output during indexing. NexusRAG and LinearRAG share the same base graph-construction pipeline, so the gap between their \texttt{Index(s)} columns ($3949.79$ versus $4188.49$\,s at 5M and $8077.13$ versus $8110.30$\,s at 10M, at most $6\%$) is run-to-run variation rather than a structural cost of the neighbor mechanism, and the comparison should be read on \texttt{Total Index (s)}.

\obs On ATLAS-Wiki, the neighbor precompute costs $106.93$\,s at 5M tokens and $228.87$\,s at 10M, i.e., $2.6\%$ and $2.8\%$ of NexusRAG's total indexing cost, so the one-time overhead of the structural propagation stays below $3\%$ at both scales. Doubling the corpus grows the total cost by $2.05\times$ for NexusRAG and $1.94\times$ for LinearRAG, essentially the same rate, while the precompute itself grows $2.14\times$, so the neighbor stage has the same scaling behavior as the base index over these two corpus sizes. Against an LLM-extraction pipeline the difference is architectural rather than incremental: HippoRAG spends $18.14 \times 10^{6}$ prompt and $9.63 \times 10^{6}$ completion tokens at 5M and $37.10 \times 10^{6}$ and $20.26 \times 10^{6}$ at 10M, whereas NexusRAG and LinearRAG consume no tokens at indexing time, and HippoRAG's total index time exceeds NexusRAG's by a factor of $17.5$ at 5M and of $12.1$ at 10M.

\end{document}